\documentclass{article}

\usepackage{PRIMEarxiv}

\usepackage[utf8]{inputenc} 
\usepackage[T1]{fontenc}    
\usepackage{hyperref}       
\usepackage{url}            
\usepackage{booktabs}       
\usepackage{amsfonts}       
\usepackage{nicefrac}       
\usepackage{microtype}      
\usepackage{lipsum}
\usepackage{fancyhdr}       
\usepackage{graphicx}       

\usepackage{pdflscape}
\usepackage{afterpage}
\usepackage{subcaption}
\usepackage{dsfont}
\usepackage[dvipsnames]{xcolor}

\usepackage{amsmath}
\usepackage{amssymb}
\usepackage{mathtools}
\usepackage{amsthm}
\usepackage{appendix}
\title{VAEneu: A New Avenue for VAE Application on Probabilistic Forecasting}

\author{
  Alireza Koochali\thanks{Corresponding author (alireza.koochali@iav.de)}\\
  IAV GmbH \\
  Kaiserslautern,Germany\\
     \And
  Ensiye Tahaei \\
  DFKI GmbH \\
  Kaiserslautern,Germany\\
  \And
  Andreas Dengel \\
  DFKI GmbH \\
  University of Kaiserslautern-Landau \\
  Kaiserslautern,Germany\\
   \And
  Sheraz Ahmed \\
  DFKI GmbH \\
  Kaiserslautern,Germany\\
}

\begin{document}
\maketitle

\begin{abstract}
This paper presents VAEneu, an innovative autoregressive method for multistep ahead univariate probabilistic time series forecasting. We employ the conditional VAE framework and optimize the lower bound of the predictive distribution likelihood function by adopting the Continuous Ranked Probability Score (CRPS), a strictly proper scoring rule, as the loss function. This novel pipeline results in forecasting sharp and well-calibrated predictive distribution. Through a comprehensive empirical study, VAEneu is rigorously benchmarked against 12 baseline models across 12 datasets. The results unequivocally demonstrate VAEneu's remarkable forecasting performance. VAEneu provides a valuable tool for quantifying future uncertainties, and our extensive empirical study lays the foundation for future comparative studies for univariate multistep ahead probabilistic forecasting.
\end{abstract}


\section{Introduction}

Probabilistic forecasting is essential in decision-making, particularly in fields where accurately assessing risk is critical, such as healthcare~\cite{avati_predicting_2018,janssoone_machine_2018,avati_improving_2018}, weather forecasting~\cite{racah_extremeweather_2017,rodrigues_deepdownscale_2018}, flood risk assessment~\cite{nevo_ml_2019}, seismic hazard prediction~\cite{mousavi_cred_2018,ross_phaselink_2019}, renewable energy sector~\cite{gensler_multi-scheme_2018}, and economic and financial risk management~\cite{huang_accurate_2018,zhang_benchmarking_2018}. These models provide insights into possible future outcomes and their likelihoods, aiding decision-makers in resource allocation, policy formulation, and strategic planning. The importance of these forecasts lies in their ability to model uncertainty about the future in the form of predictive distribution. The ultimate goal of a probabilistic forecaster is to output a predictive distribution with good calibration and sharpness. Calibration is concerned with the statistical consistency between predictive distribution and observation, while sharpness governs the confidence in the forecast itself. A sharp probabilistic forecaster outputs a narrow predictive distribution, which represents high confidence in the forecast. However, sharpness is desirable in conjunction with calibration to employ high confidence aligned with the true distribution.~\cite{gneiting_strictly_2007}

The evolution of Deep Neural Networks (DNNs) has revolutionized probabilistic forecasting, introducing methods that efficiently leverage large datasets. This advancement has shifted the focus towards end-to-end learning, minimizing the need for manual feature engineering and opening new horizons in forecasting accuracy and application. Recurrent Neural Networks (RNNs) were the first neural network architectures that were designed especially to process sequential data. The Long-Short Term Memory (LSTM)~\cite{hochreiter_long_1997} and Gated Recurrent Unit (GRU)~\cite{cho_learning_2014} are two of the most well-established RNN variants in DNN-based forecasting models. Temporal Convolutional Neural Networks (TCNs)~\cite{bai_empirical_2018} have transformed the capabilities of CNNs to time series data. Seq2Seq models~\cite{sutskever_sequence_2014}, later improved with Attention mechanism~\cite {bahdanau_neural_2016}, further enhance time series modeling.  Recently, Transformers~\cite{vaswani_attention_2017} has added to the arsenal of time series modeling tools, which further empowered the richness of the probabilistic forecasting toolbox. While the DNN models have demonstrated remarkable performance for probabilistic forecasting, they mostly restrict themselves to the family of distribution with tractable likelihood regardless of true distribution (e.g.~\cite{salinas_deepar_2020}). Another dominant approach is to model key properties of predictive distribution, such as quantile~\cite{lim_temporal_2021,wen_multi-horizon_2018} instead of modeling the predictive distribution itself. To circumvent the intractable likelihood, one newly emerging approach is to use GANs to model predictive distribution~\cite{koochali_probabilistic_2019,zhou_stock_2018}. In optimal scenarios, these models can generate samples from true data distribution. However, the instability of the adversarial objective function makes the training of these networks a daunting task. In this work, we introduced VAEneu, an autoregressive probabilistic forecaster that optimizes the lower bound of likelihood function while optimizing a strictly proper scoring rule, CRPS, to learn a sharp and well-calibrated predictive distribution without enforcing any restrictive assumption.

The main contributions of this paper are as follows:
\begin{itemize}
    \item We propose a probabilistic forecasting model based on Conditional Variational Autoencoder (CVAE), incorporating CRPS as the loss function
    \item We demonstrate the superior performance of the proposed model through extensive empirical experiments using 12 baseline models and 12 univariate datasets.
\end{itemize}

\section{Related Work}

The rise of neural networks in recent years resulted in the emergence of powerful end-to-end probabilistic forecasting methods, which are capable of effectively learning from large datasets. Khosravi et al.~\cite{khosravi_neural_2013} combined neural networks with the GARCH model~\cite{bollerslev_generalized_1986} to provide prediction interval for time series forecasting. WaveNet~\cite{oord_wavenet_2016} employed dilated convolutional neural networks (CNNs) to process long input time series data effectively and has shown outstanding performance in audio generation and time series forecasting. Long short-term memory(LSTM) networks have been used for parameterizing a linear State Space Model (SSM) to generate probabilistic forecast~\cite{rangapuram_deep_2018}. Rasp et al.~\cite{rasp_neural_2018} employed a neural network to post-process an ensemble weather prediction model. Prophet ~\cite{taylor_forecasting_2018} is a modular regression model built upon the Generalised Additive Model~\cite{hastie_generalized_1992}, which can integrate domain knowledge into the modeling process. Wen~\cite{wen_multi-horizon_2018} suggested a multi-horizon quantile forecaster (MQ-R(C)NN) based on Sequence-to-Sequence RNN (Seq2Seq)~\cite{graves_generating_2014} model. Gesthaus et al.~\cite{gasthaus_probabilistic_2019} introduced another method for quantile forecasting based on the spline functions for representing output distributions. Salinas et al.~\cite{salinas_high-dimensional_2019} proposed an autoregressive forecaster for high-dimensional multivariate time series by utilizing a low-rank covariance matrix to model the output distribution. Also, they took a copula-based approach in conjunction with an RNN model to construct the forecaster. Wang et al.~\cite{wang_deep_2019} uses a global model based on deep neural networks for capturing global non-linear patterns, while a probabilistic graphical model extracts the individual random effects locally. DeepTCN~\cite{chen_probabilistic_2020} is proposed as an encoder-decoder forecasting model based on dilated CNN models. This model can be utilized either as an explicit model or a quantile estimator. Rasul et al.~\cite{rasul_multi-variate_2020} employed conditional normalizing flows to compose an autoregressive multivariate probabilistic forecasting model. Normalizing Kalman Filter~\cite{de_bezenac_normalizing_2020} has been proposed for probabilistic forecasting, which uses normalizing flows~\cite{rezende_variational_2015} to augment a linear Gaussian state space model. DeepAR~\cite{salinas_deepar_2020} is suggested as a simple yet effective autoregressive explicit model based on RNNs. Gouttes~\cite{gouttes_probabilistic_2021} suggested an autoregressive RNN model based on Implicit Quantile Networks~\cite{dabney_implicit_2018} for quantile forecasting. Hasson et al.~\cite{hasson_probabilistic_2021} introduced Level Set Forecaster (LSF). LSF organizes training data into groups, considering those that are sufficiently similar. Subsequently, the bins containing the actual values are employed to generate the predicted distributions. Lim et al.~\cite{lim_temporal_2021} proposed the Temporal Fusion Transformer(TFT). This attention-based model utilized a sequence-to-sequence model for the local process of the input window and a temporal self-attention decoder to model long-term dependencies. Denoising diffusion models have been used to develop TimeGrad~\cite{rasul_autoregressive_2021}, an autoregressive multivariate forecaster. T{\"u}rkmen et al. ~\cite{turkmen_forecasting_2021} combined classical discrete-time conditional renewal processes~\cite{croston_forecasting_1972} with a neural networks architecture to build a forecasting model suitable for intermittent demand forecasting. Xu et al.\cite{xu_qrnn-midas_2021} expanded quantile RNN forecaster models to handle time series with mixed sample frequency. Kan et al.~\cite{kan_multivariate_2022} proposed multivariate quantile functions, which are parameterized using gradients of "input convex neural networks"~\cite{amos_input_2017}. The model employs an RNN-based feature extractor to build a probabilistic forecaster. Multiple publications~\cite{koochali_probabilistic_2019,koochali_if_2021,zhou_stock_2018,zhang_stock_2019,lin_pattern_2019} have successfully applied GAN to transform samples from a prior distribution to samples from a predictive distribution.

The recent surge in interest in long sequence time series forecasting (LSTF) in the scientific community has led to notable developments in this domain. Pioneering contributions such as Informer~\cite{zhou_informer_2021}, Autoformer~\cite{wu_autoformer_2022}, FEDformer~\cite{zhou_fedformer_2022}, and TSMixer~\cite{chen_tsmixer_2023} have been significant. These models primarily focus on the challenge of substantially extending the forecast horizon, primarily within the context of modeling multivariate time series in a deterministic manner.

\section{VAEneu}
\label{sec:vaeneu}
In this section, we delve into the specifics of our proposed methodology, which is pivotal to our study of probabilistic forecasting. Initially, we establish a formal definition and structure for the task of probabilistic forecasting, setting the stage for our subsequent discussions. We then explore the concept and mechanics of data generation using VAEs, providing a foundation for understanding our approach. Lastly, we introduce our unique contributions to this field and detail the architecture of our proposed model, highlighting how it differentiates and advances the current landscape of probabilistic forecasting models in machine learning.

\begin{figure}
    \centering
    \includegraphics[width=0.8\linewidth]{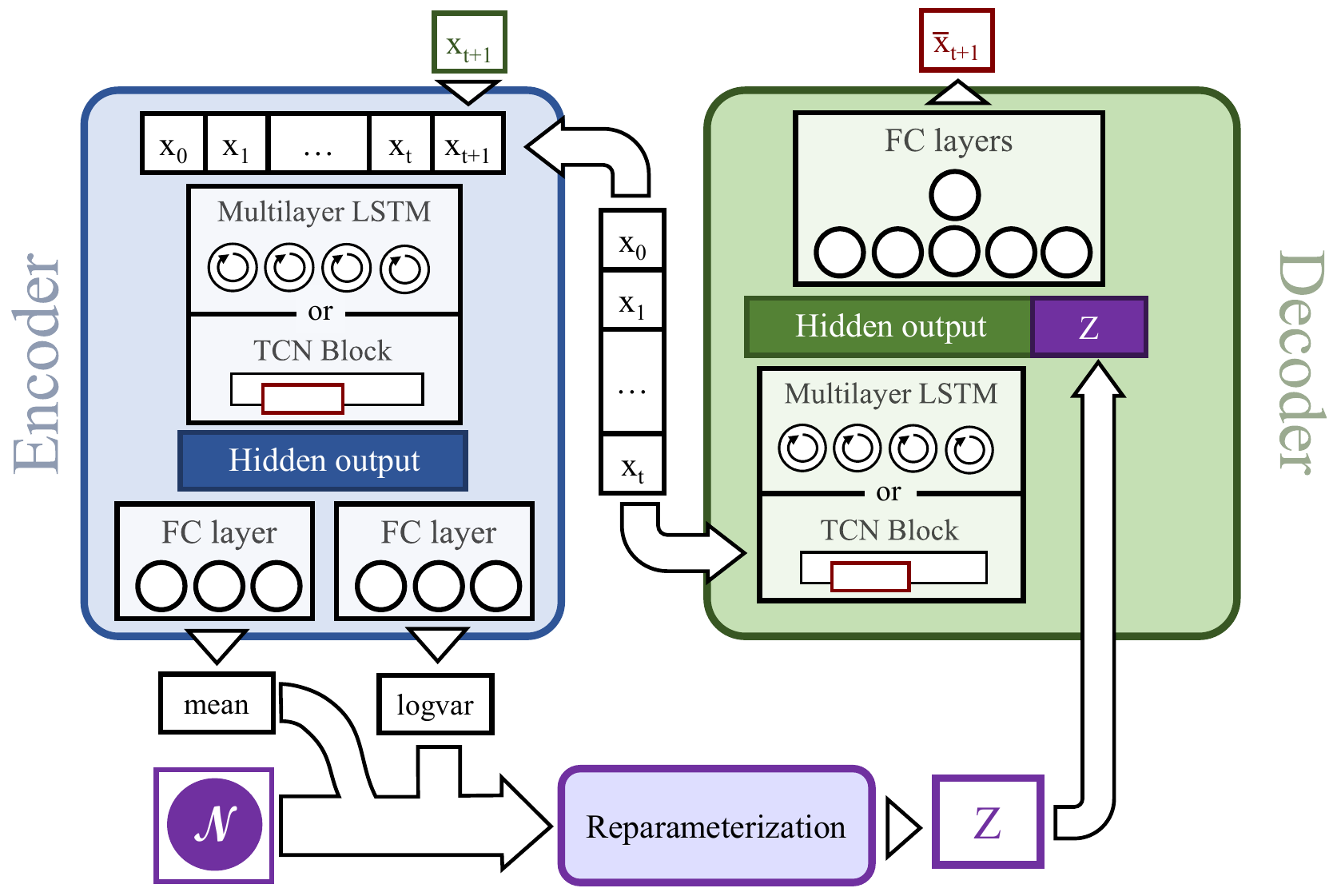}
    \caption{The structure of VAEneu in detail}
    \label{fig:vaeneu_detail}
\end{figure}
\subsection{Problem Statement}
    A given dataset consists of a set of observations generated by the process of interest. Let $\mathcal{D}$ be a dataset containing univariate time series comprising T timesteps :
    
    \begin{equation}
        \mathcal{D}=\{x_0, x_1, ..., x_T\}.
    \end{equation}
    
    In forecasting, our ultimate objective is to obtain a predictive probability distribution for future quantities based on historical information. To be more specific, we aim to model the conditional probability distribution as follows:

    \begin{equation}
    P(x_{t+1:t+h}|x_{0:t}).
    \label{eq:forecast}
    \end{equation}
    
     Here, $h$ represents the forecast horizon. 


\begin{table*}[t]
\caption{The figure displays the results of the multi-step-ahead forecasting experiment. Each model underwent three separate training iterations on each dataset. The performance is quantified using the average CRPS (lower is better), with the Coefficient of Variation (CV) indicated within parentheses. The color-coding of the cells illustrates the model's performance relative to the best-performing model on each dataset: green signifies the top model, blue indicates a deviation of up to 10\% from the best model, black for up to 50\% deviation, orange for up to 100\% deviation, and red for deviations exceeding 100\%. Notably, the VAEneu models demonstrate exemplary performance, securing the position of the best model in five datasets and, in most cases, deviating less than 50\% from the leading model's results.}

\label{tab:results}
\vskip 0.15in
\begin{center}
\begin{scriptsize}
\begin{sc}
\begin{tabular}{lcccccc}
\toprule
    &Gold price 	            & HEPC 	                & Internet traffic 	& Internet traffic & Internet traffic 	& Internet traffic 	 \\
    &            &  	                & A1H 	& A5M & B1H 	& B5M 	 \\
\midrule
Prophet& \textcolor{red}{1.1e+2 (0.3\%)}& \textcolor{black}{3.5e-1 (2.1\%)}& \textcolor{red}{4.9e+11 (1.0\%)}& \textcolor{red}{1.3e+9 (1.5\%)}& \textcolor{red}{1.9e+15 (1.7\%)}& \textcolor{red}{1.4e+14 (1.2\%)}\\
MQ-RNN& \textcolor{red}{3.2e+2 (9.4\%)}& \textcolor{red}{5.3e-1 (17.0\%)}& \textcolor{red}{1.4e+12 (0.0\%)}& \textcolor{red}{5.1e+9 (0.0\%)}& \textcolor{red}{1.6e+16 (0.0\%)}& \textcolor{red}{1.2e+15 (0.0\%)}\\
MQ-CNN& \textcolor{red}{4.6e+1 (15.5\%)}& \textcolor{red}{7.3e-1 (27.7\%)}& \textcolor{red}{4.7e+11 (3.8\%)}& \textcolor{red}{9.3e+8 (1.6\%)}& \textcolor{red}{2.4e+15 (8.1\%)}& \textcolor{red}{2.4e+14 (3.1\%)}\\
TFT& \textcolor{red}{4.7e+1 (18.4\%)}& \textcolor{black}{2.9e-1 (24.5\%)}& \textcolor{red}{2.5e+11 (42.6\%)}& \textcolor{orange}{1.7e+8 (5.3\%)}& \textcolor{orange}{8.1e+14 (3.0\%)}& \textbf{\textcolor{Green}{3.0e+13 (10.1\%)}}\\
Transformer& \textcolor{red}{7.8e+1 (55.8\%)}& \textcolor{NavyBlue}{2.8e-1 (7.8\%)}& \textcolor{orange}{1.0e+11 (37.4\%)}& \textcolor{orange}{1.8e+8 (32.5\%)}& \textcolor{orange}{8.6e+14 (19.2\%)}& \textcolor{orange}{5.1e+13 (31.7\%)}\\
DeepFactor& \textcolor{red}{1.1e+2 (77.4\%)}& \textcolor{red}{1.1e+0 (9.9\%)}& \textcolor{red}{1.5e+12 (55.5\%)}& \textcolor{red}{3.4e+9 (2.2\%)}& \textcolor{red}{5.8e+15 (24.3\%)}& \textcolor{red}{4.9e+14 (4.1\%)}\\
DeepAR& \textcolor{red}{6.5e+1 (48.2\%)}& \textbf{\textcolor{Green}{2.6e-1 (4.2\%)}}& \textcolor{orange}{9.9e+10 (11.4\%)}& \textcolor{NavyBlue}{1.0e+8 (13.4\%)}& \textcolor{black}{6.5e+14 (29.3\%)}& \textcolor{black}{3.5e+13 (5.5\%)}\\
DRP& \textcolor{red}{1.3e+3 (0.2\%)}& \textcolor{red}{5.6e-1 (2.4\%)}& \textcolor{red}{1.4e+12 (0.0\%)}& \textcolor{red}{5.2e+9 (0.0\%)}& \textcolor{red}{1.6e+16 (0.0\%)}& \textcolor{red}{1.2e+15 (0.0\%)}\\
Wavenet& \textcolor{orange}{3.3e+1 (23.2\%)}& \textcolor{black}{2.9e-1 (6.7\%)}& \textcolor{red}{1.8e+11 (23.9\%)}& \textcolor{black}{1.3e+8 (9.3\%)}& \textcolor{red}{1.5e+15 (14.2\%)}& \textcolor{NavyBlue}{3.3e+13 (22.6\%)}\\
GPForecaster& \textcolor{red}{3.9e+1 (2.2\%)}& \textcolor{red}{6.0e-1 (1.5\%)}& \textcolor{red}{1.4e+12 (0.0\%)}& \textcolor{red}{5.1e+9 (0.0\%)}& \textcolor{red}{1.6e+16 (0.0\%)}& \textcolor{red}{1.2e+15 (0.0\%)}\\
DeepState& \textcolor{red}{4.5e+1 (10.3\%)}& \textcolor{red}{7.2e-1 (12.4\%)}& \textcolor{red}{7.9e+11 (5.2\%)}& \textcolor{red}{1.2e+9 (25.1\%)}& \textcolor{red}{2.8e+15 (8.6\%)}& \textcolor{red}{2.7e+14 (19.7\%)}\\
ForGAN& \textcolor{NavyBlue}{1.9e+1 (4.2\%)}& \textcolor{black}{3.0e-1 (5.7\%)}& \textcolor{orange}{8.9e+10 (4.6\%)}& \textbf{\textcolor{Green}{9.6e+7 (4.6\%)}}& \textcolor{red}{1.1e+15 (35.1\%)}& \textcolor{black}{4.0e+13 (6.0\%)}\\
\midrule
VAEneu-RNN& \textbf{\textcolor{Green}{1.9e+1 (1.3\%)}}& \textcolor{black}{3.1e-1 (8.4\%)}& \textbf{\textcolor{Green}{5.3e+10 (5.8\%)}}& \textcolor{black}{1.1e+8 (6.1\%)}& \textcolor{NavyBlue}{5.8e+14 (20.7\%)}& \textcolor{black}{4.2e+13 (12.6\%)}\\
VAEneu-TCN& \textcolor{black}{2.1e+1 (2.7\%)}& \textcolor{black}{2.9e-1 (3.7\%)}& \textcolor{black}{5.9e+10 (4.0\%)}& \textcolor{black}{1.1e+8 (12.7\%)}& \textbf{\textcolor{Green}{5.3e+14 (11.4\%)}}& \textcolor{black}{3.5e+13 (7.0\%)}\\

\bottomrule
\\
\toprule
    & Mackey Glass 	        & Saugeen river 	 & Solar 4 seconds 	    & Sunspot 	            & US births 	            & Wind 4 seconds 	 \\
\midrule
Prophet& \textcolor{red}{1.3e-1 (0.7\%)}& \textcolor{black}{1.2e+1 (2.3\%)}& \textcolor{red}{7.5e+3 (1.4\%)}& \textcolor{red}{2.2e+1 (1.5\%)}& \textcolor{red}{3.6e+2 (0.9\%)}& \textcolor{orange}{1.2e+4 (2.4\%)}\\
MQ-RNN& \textcolor{red}{1.4e-1 (5.2\%)}& \textcolor{red}{2.1e+1 (3.4\%)}& \textcolor{red}{5.4e+3 (8.1\%)}& \textcolor{black}{3.4e+0 (4.8\%)}& \textcolor{red}{9.5e+2 (2.4\%)}& \textcolor{red}{1.4e+4 (1.3\%)}\\
MQ-CNN& \textcolor{red}{1.2e-1 (7.4\%)}& \textcolor{orange}{1.4e+1 (1.2\%)}& \textcolor{red}{4.8e+3 (11.9\%)}& \textcolor{black}{3.5e+0 (14.6\%)}& \textcolor{red}{5.0e+2 (14.2\%)}& \textcolor{orange}{1.3e+4 (1.5\%)}\\
TFT& \textcolor{red}{5.9e-3 (17.8\%)}& \textcolor{NavyBlue}{8.9e+0 (7.0\%)}& \textcolor{red}{3.3e+3 (18.2\%)}& \textcolor{black}{3.0e+0 (3.0\%)}& \textbf{\textcolor{Green}{1.3e+2 (1.8\%)}}& \textcolor{black}{9.3e+3 (14.0\%)}\\
Transformer& \textcolor{red}{1.2e-1 (19.1\%)}& \textcolor{black}{1.2e+1 (14.9\%)}& \textcolor{black}{1.9e+3 (10.4\%)}& \textcolor{NavyBlue}{2.6e+0 (0.7\%)}& \textcolor{red}{3.9e+2 (58.9\%)}& \textcolor{black}{8.9e+3 (23.1\%)}\\
DeepFactor& \textcolor{red}{2.5e-1 (1.9\%)}& \textcolor{red}{2.2e+1 (1.0\%)}& \textcolor{red}{3.0e+4 (2.4\%)}& \textcolor{NavyBlue}{2.8e+0 (12.3\%)}& \textcolor{red}{3.1e+3 (53.2\%)}& \textcolor{orange}{1.2e+4 (3.9\%)}\\
DeepAR& \textcolor{red}{7.6e-2 (8.1\%)}& \textcolor{orange}{1.3e+1 (2.4\%)}& \textbf{\textcolor{Green}{1.6e+3 (13.3\%)}}& \textcolor{NavyBlue}{2.6e+0 (2.3\%)}& \textcolor{black}{1.8e+2 (7.8\%)}& \textcolor{black}{9.7e+3 (12.0\%)}\\
DRP& \textcolor{red}{1.9e-1 (0.0\%)}& \textcolor{black}{1.0e+1 (2.4\%)}& \textcolor{red}{3.0e+4 (0.0\%)}& \textcolor{red}{3.4e+1 (23.4\%)}& \textcolor{red}{1.1e+4 (0.1\%)}& \textcolor{red}{1.7e+4 (0.0\%)}\\
Wavenet& \textcolor{red}{7.1e-3 (1.8\%)}& \textcolor{black}{9.4e+0 (4.7\%)}& \textcolor{red}{4.2e+3 (31.7\%)}& \textbf{\textcolor{Green}{2.6e+0 (12.0\%)}}& \textcolor{red}{3.2e+2 (10.6\%)}& \textcolor{orange}{1.3e+4 (22.0\%)}\\
GPForecaster& \textcolor{red}{3.3e-1 (44.3\%)}& \textcolor{black}{1.1e+1 (1.8\%)}& \textcolor{red}{2.1e+4 (1.6\%)}& \textcolor{orange}{4.0e+0 (16.5\%)}& \textcolor{red}{4.2e+2 (0.2\%)}& \textcolor{orange}{1.2e+4 (3.1\%)}\\
DeepState& \textcolor{red}{2.8e-1 (1.1\%)}& \textcolor{orange}{1.5e+1 (13.4\%)}& \textcolor{red}{7.4e+3 (18.3\%)}& \textcolor{red}{9.9e+0 (13.7\%)}& \textcolor{red}{5.4e+2 (4.8\%)}& \textcolor{red}{2.8e+4 (28.3\%)}\\
ForGAN& \textcolor{red}{4.9e-3 (31.2\%)}& \textbf{\textcolor{Green}{8.3e+0 (0.5\%)}}& \textcolor{orange}{3.1e+3 (6.4\%)}& \textcolor{black}{3.6e+0 (3.4\%)}& \textcolor{red}{2.8e+2 (6.8\%)}& \textcolor{NavyBlue}{6.9e+3 (3.6\%)}\\ \midrule
VAEneu-RNN& \textcolor{orange}{8.6e-4 (10.8\%)}& \textcolor{NavyBlue}{8.4e+0 (0.2\%)}& \textcolor{orange}{2.8e+3 (3.7\%)}& \textcolor{black}{3.4e+0 (2.4\%)}& \textcolor{red}{2.9e+2 (3.4\%)}& \textbf{\textcolor{Green}{6.9e+3 (0.3\%)}}\\
VAEneu-TCN& \textbf{\textcolor{Green}{4.5e-4 (24.9\%)}}& \textcolor{NavyBlue}{8.4e+0 (1.1\%)}& \textcolor{black}{1.8e+3 (7.5\%)}& \textcolor{NavyBlue}{2.7e+0 (1.5\%)}& \textcolor{red}{2.9e+2 (7.1\%)}& \textcolor{NavyBlue}{7.4e+3 (0.5\%)}\\

\bottomrule
\end{tabular}
\end{sc}
\end{scriptsize}
\end{center}
\vskip -0.1in
\end{table*}

\subsection{Variational Autoencoders}

Variational Autoencoder (VAE)~\cite{kingma_auto-encoding_2013} is a member of explicit generative models and latent viable models. The aim of the model is to generate artificial data x from the generative distribution \(p_\theta(x|z)\) conditioned on latent variable \(z\sim p_\theta(z)\). Due to intractable true posterior \(p_\theta(z|x)\), the direct estimation of the parameter \(\theta\) is not feasible. However, VAE defines a surrogate function \(q_\phi(z|x)\), which is also known as the recognition model, to approximate the true posterior. With this initiative, the VAE can efficiently be trained via stochastic gradient decent using the variational lower bound of the log-likelihood:

    \begin{equation}
    \begin{split}
\mathcal{L}_{\mathrm{VAE}}(\mathbf{x} ; \theta, \phi)= &-KL\left(q_\phi(\mathbf{z} \mid \mathbf{x}) \| p_\theta(\mathbf{z})\right) \\
& +\mathbb{E}_{q_\phi(\mathbf{z} \mid \mathbf{x})}\left[\log p_\theta(\mathbf{x} \mid \mathbf{z})\right].
    \end{split}
    \end{equation}

Assuming the Gaussian latent variable, the first term of the loss function can be integrated analytically. The second term, which is also known as reconstruction error, can be approximated by drawing multiple samples \(z^{(l)} (l = 1, ..., L)\) from \(q_\phi(z|x)\); hence the empirical loss function of VAE with Gaussian latent variable is:

\begin{equation}
\begin{split}
\widetilde{\mathcal{L}}_{\mathrm{VAE}}(\mathbf{x} ; \theta, \phi)= &-K L\left(q_\phi(\mathbf{z} \mid \mathbf{x}) \| p_\theta(\mathbf{z})\right) \\ & +\frac{1}{L} \sum_{l=1}^L \log p_\theta\left(\mathbf{x} \mid \mathbf{z}^{(l)}\right)
\end{split}
\end{equation}

To decrease the variance of the gradients for training, VAE uses the reparameterization trick where the recognition distribution \(q_\phi(z|x)\) is parameterized via a deterministic, differentiable function \(g_\phi(.,.)\); thus, $
\mathbf{z}^{(l)}=g_\phi\left(\mathbf{x}, \epsilon^{(l)}\right), \epsilon^{(l)} \sim \mathcal{N}(\mathbf{0}, \mathbf{I})
$.

In practice, the prior over latent variable is considered as an isotropic multivariate Gaussian \(p_\theta(z)=\mathcal{N}(z;0;\mathbf{I})\). The \(q_\phi(z|x)\) and \(p_\theta(x|z)\) are approximated with neural networks, namely encoder and decoder, respectively. Furthermore, the \(q_\phi(z|x)\) is parameterized using \(g_\phi(x,\epsilon^{(l)}) = \mu + \sigma \odot \epsilon^{(l)}\) where x is data point and \(\odot\) signifies the element-wise product.

Conditional VAE (CVAE)~\cite{sohn_learning_2015} is an important extension of VAEs that enables us to train VAE conditioned on some auxiliary information \(c\) and generate artificial data from \(p_\theta(x|c)\). To do so, the condition is presented to both encoder and decoder and is integrated into VAE's objective function:

\begin{equation}
\begin{split}
\widetilde{\mathcal{L}}_{\mathrm{CVAE}}(\mathbf{x},\mathbf{c} ; \theta, \phi)= &-K L\left(q_\phi(\mathbf{z} \mid \mathbf{x},\mathbf{c}) \| p_\theta(\mathbf{z})\right) \\ & +\frac{1}{L} \sum_{l=1}^L \log p_\theta\left(\mathbf{x} \mid \mathbf{z}^{(l)},\mathbf{c}\right)
\end{split}
\end{equation} 

As stated in Equation~\ref{eq:forecast}, the objective of probabilistic forecasting is to learn the distribution of future values, also known as predictive distribution, conditioned on historical window input. Since CVAE provides us with a tool to learn conditional probability distribution, with the right setup, we can train it as a probabilistic forecaster. To do so, we need to use historical window input as the condition, i.e., \(c = x_{0:t}\), and the target future values as the CVAE input data.

In this work, we aim to model the one-step ahead forecasting to simplify the training pipeline. During inference, we extend the resultant model into the multi-step forecasting model using auto-regression, i.e., feeding the model's forecast back to the model to forecast further in the future. Hence, the predictive distribution is updated as \(p_\theta(x_{t+1}|z,x_{x:0})\). 

To train CVAE as the probabilistic forecaster, we still need to define a loss function that lets the model learn the intrinsic and complexity of predictive distribution. The conventional approach in VAE is that when the data is a continuous value, the likelihood function is assumed to be multivariate Gaussian, i.e., \(p_\theta(x|z) = \mathcal{N}(x;\mu_\theta(z),\Sigma_\theta(z))\). To streamline the training process, the \(\Sigma_\theta(z)\) is assumed to be a diagonal matrix which can be written as \(\Sigma_\theta(z) = \sigma\mathbf{I}\). This assumption effectively turns the covariance of predictive distribution as constant and simplifies the reconstruction error to the mean squared error (MSE). These assumptions and simplification enable the efficient training of VAE on continuous variables. However, probabilistic forecasters trained using these assumptions tend to generate a very sharp predictive distribution around the mean of the predictive distribution and disregard statistical constancy with true data distribution, which results in low calibration. Therefore, we need a novel reconstruction loss that enforces diversity between samples and lets the model learn the nuances of the predictive distribution.

\begin{figure*}
    \centering
    \includegraphics[width=\linewidth]{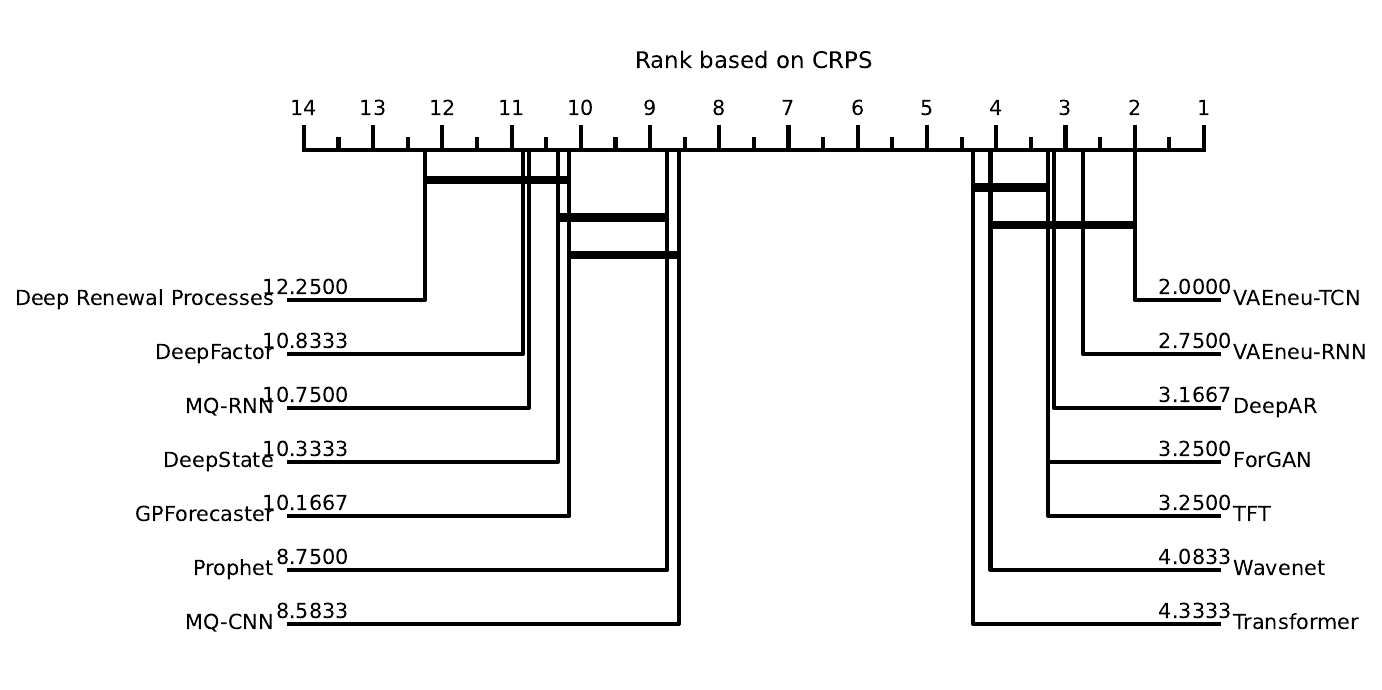}
    \caption{The critical difference diagram for univariate time series probabilistic forecasting experiment in which two separate clusters of models emerge based on
    the performance of them on all datasets.}
    \label{fig:cd}
\end{figure*}

\subsection{CRPS Estimation as Reconstruction Loss}
The Continuous Ranked Probability Score (CRPS) serves as a univariate, strictly proper scoring rule, evaluating the alignment of a cumulative distribution function (CDF) $F$ with an observed value $x \in \mathbb{R}$. This is formally expressed as:
\begin{equation}
    \operatorname{CRPS}(F, x)=\int_{\mathbb{R}}(F(y)-\mathds{1}\{x \leq y\})^{2} dy \,,
    \label{eq:crps_F}
\end{equation}
\noindent
where $\mathds{1}\{x \leq y\}$ denotes the indicator function, taking a value of one if $x \leq y$ and zero otherwise. Since  CRPS is a strictly proper scoring rule, it reaches its maximum when the predictive distribution precisely mirrors the actual data distribution. Consequently, minimizing the negative CRPS as a reconstruction loss for training a CVAE-based probabilistic forecaster effectively equates to optimizing the likelihood function of the predictive distribution, thereby making CRPS an effective proxy for optimizing \(p_\theta(x_{t+1}|z,x_{0:t})\) in the CVAE pipeline.

On the other hand, during inference from a CVAE, direct access to the CDF of the predictive distribution is not available. Instead, the model provides samples from the predictive distribution. There is a method to approximate the CRPS in this scenario. Using lemma 2.2 of~\cite{baringhaus_new_2004} or identity 17 of~\cite{szekely_new_2005}, we can define the negative orientation of $\operatorname{CRPS}$ by
\begin{equation}
    \operatorname{CRPS}(F, x)=\mathrm{E}_{F}\left|\bar{X}-x\right|-\frac{1}{2} \mathrm{E}_{F}\left|\bar{X}-\bar{X}^{\prime}\right|\,,
    \label{eq:crps}
\end{equation}
\noindent
where $\bar{X}$ and $\bar{X}^{\prime}$ are independent copies of a random variable
with distribution function F and a finite first moment~\cite{gneiting_strictly_2007}. In practice, we calculate CRPS using Equation~\ref{eq:crps} by drawing two sets of samples from the predictive distribution to represent $\bar{X}$ and $\bar{X}^{\prime}$. The accuracy of CRPS estimation is dependent on the sample size of $\bar{X}$ and $\bar{X}^{\prime}$.

Intuitively, optimizing the CRPS, as defined in Equation~\ref{eq:crps}, involves balancing two contrasting objectives. The first term, representing the mean absolute error between the forecast and the target, aims to sharpen the predictive distribution around the median. Simultaneously, the second term, assessing the spread of forecast samples, encourages diversity in forecast samples to enhance calibration. The optimization process thus navigates towards a balanced predictive distribution, one that reconciles sharpness with a well-calibrated spread of forecast samples.

\subsection{Proposed Model}

In this study, we optimized the CRPS approximation (Equation~\ref{eq:crps}) to train a CVAE for probabilistic forecasting. For a given input \(x_{t+1}\) and condition \(x_{t:0}\), we sampled multiple instances \(z\) from the recognition distribution \(q_\theta(z|x_{t:0}, x_{t+1})\) to obtain forecast samples for $\bar{X}$. The number of these samples, denoted as \textit{sample size}, is a critical hyperparameter. To compute CRPS efficiently, we shuffled the forecast samples in $\bar{X}$ to create $\bar{X}^{\prime}$, thereby circumventing the need for additional sampling from the predictive distribution.

\begin{figure*}
     \captionsetup[subfigure]{labelformat=empty}
     \centering
    \begin{subfigure}{0.32\textwidth}
         \centering
         \includegraphics[width=\textwidth]{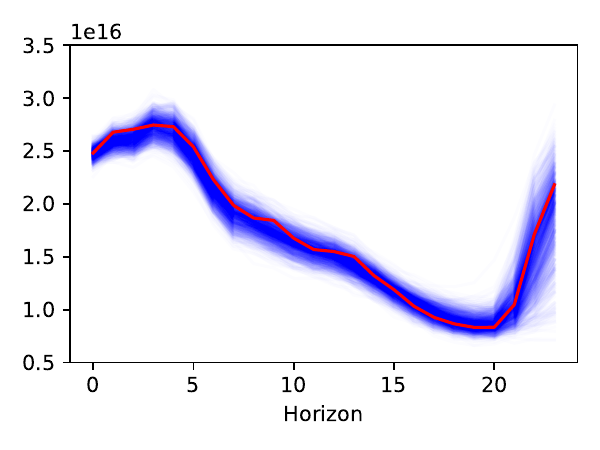}
         \caption{VAEneu-TCN}
     \end{subfigure}
     \hfill
     \begin{subfigure}{0.32\textwidth}
         \centering
         \includegraphics[width=\textwidth]{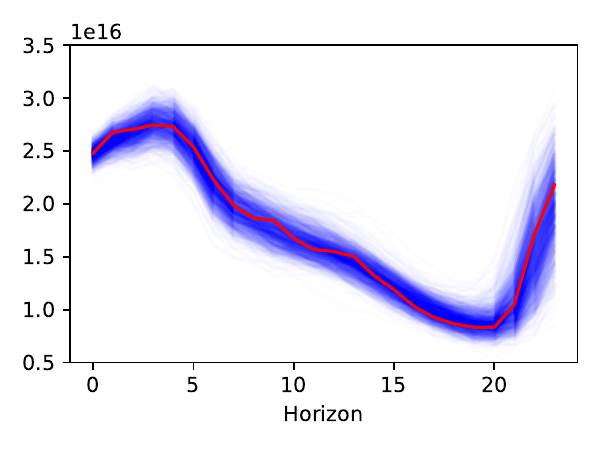}
         \caption{VAEneu-RNN}
     \end{subfigure}
     \hfill
    \begin{subfigure}{0.32\textwidth}
         \centering
         \includegraphics[width=\textwidth]{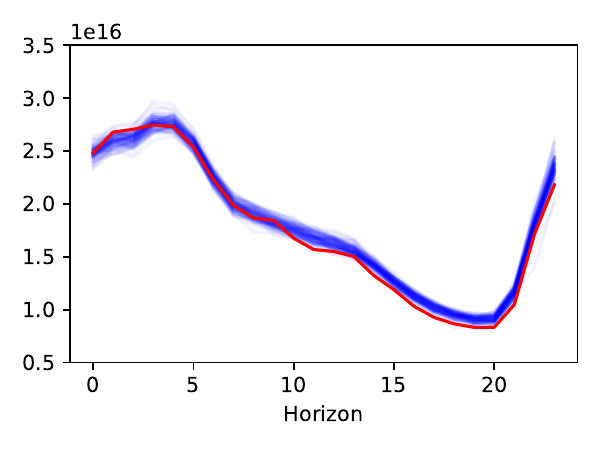}
         \caption{DeepAR (top baseline)}
     \end{subfigure}
     \vspace*{-3mm} 
        \caption{The qualitative results of VAEneu models alongside the top baseline model, DeepAR, on the Internet Traffic B1h dataset demonstrate that the proposed models' calibrations are relatively superior in comparison to DeepAR predictive distribution.}
        \label{fig:sample}
\end{figure*}

Figure~\ref{fig:vaeneu_detail} offers a detailed view of the VAEneu architecture, which was implemented in two distinct structural variants. VAEneu-TCN incorporates TCN in both its encoder and decoder, while VAEneu-RNN employs LSTM networks as its core components. The encoder process involves concatenating the historical data with the target forecast to form the network input \(x_{0:t+1}\). This input is then transformed into a representation \(h_{0:t+1}\) using either LSTM or TCN. Subsequently, the representation passes through parallel fully connected layers to generate mean \(\mu_\phi(h_{0:t+1})\) and standard deviation \(\sigma_\phi(h_{0:t+1})\) outputs. Latent variables \(z\) are generated via reparameterization and then fed into the decoder.

The decoder's role is to learn the predictive distribution \(p_\theta(x_{t+1}|z,x_{0:t})\). It begins by mapping the condition to its representation \(h_{0:t}\) and then concatenates this with the latent variable from the encoder. The final mapping to the next forecast is achieved through a fully connected layer.

\section{Experiment}
\subsection{Datasets}

This section introduces an array of datasets utilized to evaluate VAEneu performance nuances under different data conditions. We employed 12 datasets sourced from public repositories, namely Gold Price, Household Electric Power Consumption(HEPC), Internet Traffic Datasets(A5m, A1h, B5m, B1h), Macky-Glass, Saugeen River Flow, Solar 4 second, Wind 4 second, Sunspot and US birth. Appendix~\ref{app:ds_rev} presents salient features of these datasets alongside a detailed review of each dataset and a visual representation of them.

\subsection{Baseline}
In this study, we compare the proposed model with 12 established probabilistic forecasting models namely DeepAR~\cite{salinas_deepar_2020}, DeepState~\cite{rangapuram_deep_2018}, DeepFactor~\cite{wang_deep_2019}, Deep Renewal Processes (DRP)~\cite{turkmen_forecasting_2021}, GPForecaster, MQ-RNN and MQ-CNN~\cite{wen_multi-horizon_2018}, Prophet~\cite{taylor_forecasting_2018}, Wavenet~\cite{oord_wavenet_2016}, Transformer~\cite{vaswani_attention_2017}, TFT~\cite{lim_temporal_2021}, ForGAN~\cite{koochali_probabilistic_2019}.
TFT, MQ-RNN, and MQ-CNN models provide quantiles of predictive distribution, while the rest of the models generate samples from predictive distribution as the forecast. We employed the implementation of these methods from GluonTS~\cite{alexandrov_gluonts_2020} package with their default hyperparameters, with the exception of ForGAN. For the ForGAN, we used the implementation provided by their author. Appendix~\ref{app:baseline} provides detailed information on baseline models.

\subsection{Assessment}

In this study, we utilize the CRPS to assess the performance of various probabilistic forecasters. For models that yield samples from the predictive distribution as their forecast output, we used 1,000 samples per time step in the forecast horizon from predictive distribution to approximate CRPS using Equation~\ref{eq:crps}.
For probabilistic forecasting models that provide quantiles of the predictive distribution for their forecasts, we engage a quantile-based approximation method for CRPS calculation.

CRPS is conceptually interpreted as the integrated pinball loss across all possible quantile levels $\alpha$, ranging from 0 to 1~\cite{gasthaus_probabilistic_2019}. This relationship is mathematically expressed as:
\begin{equation}
\operatorname{CRPS}\left(F^{-1}, x\right)=\int_{0}^{1} 2 \Lambda_{\alpha}\left(F^{-1}(\alpha), x\right) d \alpha,,
\label{eq:crps_pinball}
\end{equation}

where $F^{-1}$ signifies the quantile function. For practical applications, the quantile-based CRPS is typically approximated based on the available set of quantiles. Thus, Equation~\ref{eq:crps_pinball} is effectively approximated as a summation over a specified number of quantiles, N. The accuracy of this approximation is dependent upon the number of quantiles utilized, influencing the precision of the CRPS estimation. We employed 99 equally distant quantile levels to estimate CRPS.

\subsection{Experiment Setup}

Our experimental setup involves partitioning each dataset, reserving a portion equivalent to five forecast horizons for testing while the remainder is utilized for model training. The proposed models undergo training using RMSProp optimizer~\cite{tieleman_lecture_2012} for a maximum of 100,000 steps. However, training is ceased prematurely if no improvement in model performance is observed over 5,000 consecutive training steps.

The models detailed in this study have been implemented leveraging the Pytorch framework~\cite{paszke_pytorch_2019} and trained on a machine equipped with an Nvidia RTX 6000 GPU. We used eight samples to estimate CRPS during training (sample size = 8). The rest of the hyperparameters for VAEnue models are presented in Appendix~\ref{app:hyp}.

 To circumvent any anomalies and ensure robustness, each model is subjected to three independent training iterations on every dataset. The reported performance metric is an arithmetic mean of these three runs, which is indicated as $\overline{CRPS}$. The coefficient of variation (CV) is also presented in parentheses to shed light on the stability and reproducibility of each technique. CV provides a standardized measure of dispersion of a probability distribution or frequency distribution and is defined as:
\begin{equation}
    CV = \frac{\sigma}{\mu} \times 100\%
\end{equation}
A lower CV indicates less variability relative to the mean.

\begin{figure}
    \centering
    \includegraphics[width=0.6\linewidth]{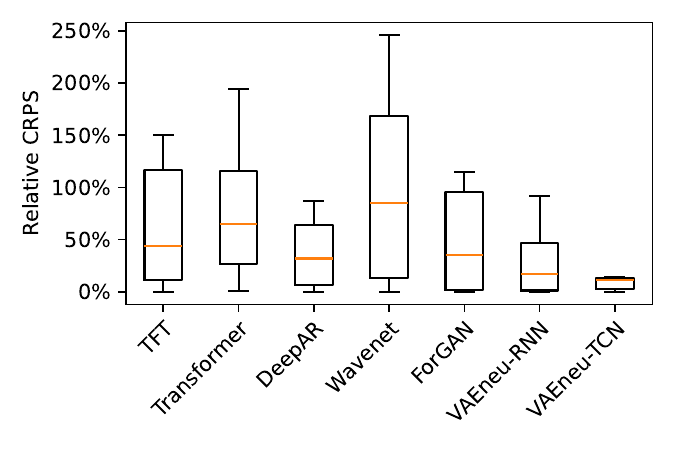}
    \caption{The distribution of relative CRPS for top tier models across all datasets revealing remarkable performance of VAEneu-TCN model which is constantly secured best results or performed close to the best model on all datasets.}
    \label{fig:rel}
\end{figure}

\section{Results and Discussion}

The comprehensive evaluation of VAEneu models and 12 baselines across 12 diverse datasets is encapsulated in Table~\ref{tab:results}. The extensive data presented in these tables can initially seem overwhelming. However, a color-coded scheme aids in interpreting the relative performance of the models. The color denotes the relative score of a model in comparison to the best-performing model on that dataset and it is defined as:

\begin{equation}
    \Delta_{CRPS} = \frac{\overline{CRPS} - \overline{CRPS}^*}{\overline{CRPS}^*} \times 100\%,
\end{equation}

wherein $\overline{CRPS}^*$ represents the $\overline{CRPS}$ of the top-performing model for a given dataset. With a glance at the color scheme of Table~\ref{tab:results}, we can perceive that VAEneu models are the best-performing model on five datasets, and on the rest of the datasets, they are always following the best-performing model closely, with the expectation of one dataset, US Birth dataset.
\begin{figure}
    \centering
    \includegraphics[width=0.6\linewidth]{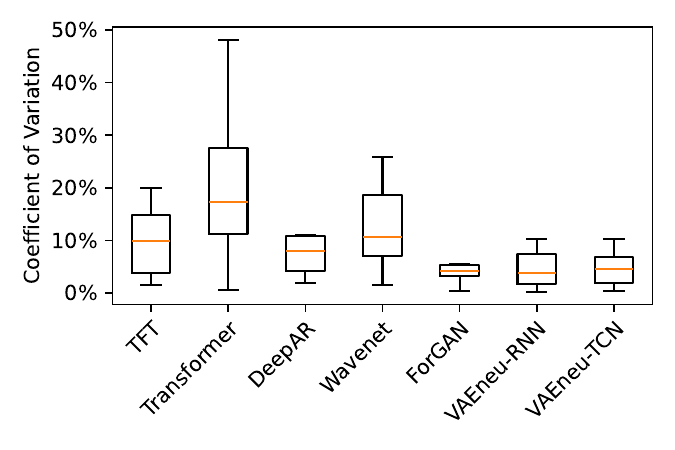}

    \caption{The coefficient of variation distribution for top tier models across all datasets indicates the consistent performance of VAEneu models across three training iteration overall datasets and highlights stable performance for the ForGAN and DeepAR models as well.}
    \label{fig:cv}
\end{figure}

\begin{figure*}
     \captionsetup[subfigure]{labelformat=empty}
     \centering
    \begin{subfigure}{0.32\textwidth}
         \centering
         \includegraphics[width=0.9\textwidth]{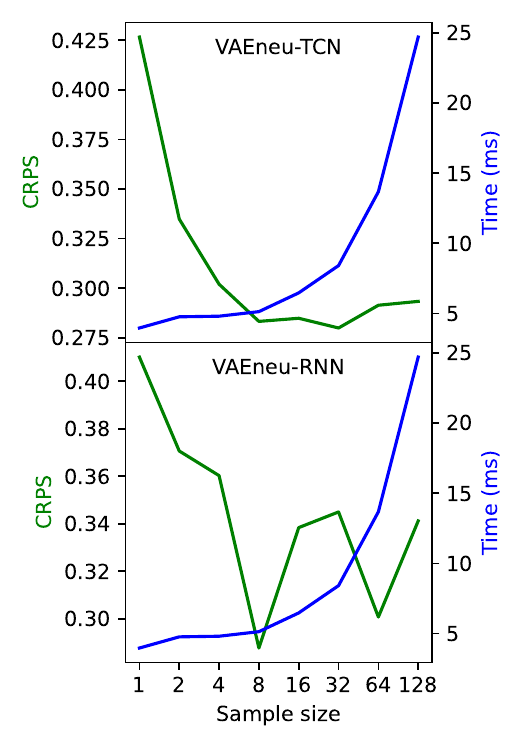}
         \caption{HEPC Dataset}
     \end{subfigure}
     \hfill
     \begin{subfigure}{0.32\textwidth}
         \centering
         \includegraphics[width=0.9\textwidth]{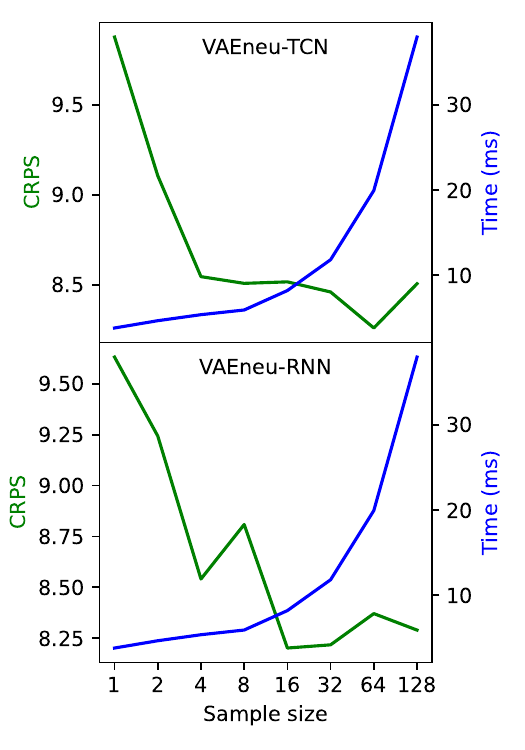}
         \caption{Saugeenday River Dataset}
     \end{subfigure}
     \hfill
    \begin{subfigure}{0.32\textwidth}
         \centering
         \includegraphics[width=0.9\textwidth]{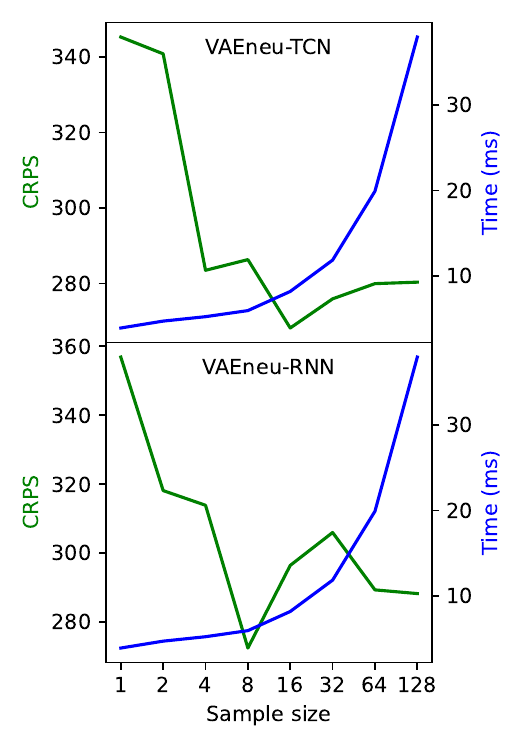}
         \caption{US Birth Dataset}
     \end{subfigure}
        \caption{The effect of sample size on the CRPS of final model and training step processing time on three datasets. A larger sample size would result in a better model in exchange for a longer training time; however, with sample size\(\in [4,32]\), a model can reach a low CRPS without increasing the batch processing time significantly. }
        \label{fig:repeat_factor}
\end{figure*}

To synthesize these extensive results into a more digestible format, we utilize the critical difference diagram (CD diagram)~\cite{ismail_fawaz_deep_2019}. The CD diagram reflects on the rank of models on each dataset and paints a complete picture of models performance on all datasets. Figure~\ref{fig:cd} showcases this diagram for the multi-step-ahead forecasting experiment. The model's names are on the side with the average rank across all datasets by their side. The thick horizontal lines are called critical difference bands. The performance of the models, which are grouped by a critical difference band, is not significantly different considering $\overline{CRPS}$ of them across all datasets. The critical difference bands are defined using the Wilcoxon signed-rank test at a significance threshold of 0.05, and the model ranking is done via the one-sided paired t-test at a 0.05 significant threshold. Appendix~\ref{app:cd} provides a complete description of how the CD diagram has been created for this study. From this figure, we can clearly perceive that the models fall into two categories. The top critical difference band includes VAEneu models alongside DeepAR, Forgan, TFT, Wavenet, and Transformer, suggesting their comparable performance across all datasets. These models, hereafter referred to as "top-tier models," exhibit performances that are statistically superior compared to the rest of the baseline models.

Next, we delved deeper into the performance analysis of top-tier models. Figure~\ref{fig:rel} illustrates the distribution of relative scores \(\Delta_{CRPS}\) among the top-tier models across all datasets. VAEneu-TCN separates itself from the rest of the models with a stable performance that is perfect or near perfect across all datasets. VAEneu-RNN follows the VAEnet-TCN alongside DeepAR and ForGAN. DeepAR's remarkable performance was somewhat unexpected, considering that DeepAR assumes explicitly that predictive distribution is Gaussian distribution, which renders the model less flexible. The performance of DeepAR suggests that Gaussian distributions can aptly approximate uncertainty in many real-world scenarios. 

Figure~\ref{fig:sample} exhibits VAEneu models forecast on the Internet Traffic B1H dataset and provides qualitative comparison ground against DeepAR, the best-performing baseline model on this dataset. The DeepAR model generates a sharp and relatively low-calibrated predictive distribution. In contrast, the VAEneu models demonstrate an effective balance between sharpness and calibration within their predictive distributions. Notably, the TCN-based model has slightly better sharpness in comparison to the RNN variant.

Finally, to analyze the model performance consistency across three runs, we investigate the distribution of CV for top-tier models and present the result in Figure~\ref{fig:cv}. Here, we can observe that VAEneu, alongside ForGAN and DeepAR models, can deliver good performance at a consistent rate. Appendix~\ref{app:qr} illustrates forecasts from VAEneu models and baseline to provide a qualitative insight into the models' performances as well. 

\section{Analyzing the Impact of the Sample Size on Training}

The \emph{sample size} designates the number of samples for each input used to approximate the CRPS during training. A larger sample size enhances the accuracy of CRPS estimation. This, in turn, should refine the quality of gradients obtained from the CRPS loss, facilitating the weight update process within the network. Nevertheless, the influence of the sample size on the model's training trajectory remains an area yet to be fully explored. In this section, we aim to clarify the consequence of varying sample size values on both the CRPS of the resultant optimal model and the average time taken to complete a training step.

To assess the interplay of different sample sizes, we trained VAEneu architectures across three distinct datasets. Sample sizes were chosen following the relation:
\begin{equation}
    \text{sample size} = 2^i, \text{where}~ i \in \{0,1, 2, \ldots, 7\}.
\end{equation}

Figure~\ref{fig:repeat_factor} encapsulates our findings. An evident trend emerges: as the sample size escalates, models tend to converge to a configuration with a more favorable CRPS. This improvement, however, comes at the cost of an increased batch processing duration. In the case where the \(\text{sample size} = 1\), the CRPS loss turns into MAE loss. From this figure, we can observe that moving from MAE loss toward CRPS loss improves the model performance significantly, underlying the effectiveness of the proposed methodology. Moreover, an interesting observation to note is that by choosing a sample size within the interval $[4,32]$, one can achieve decent CRPS values without significantly extending the training step processing time.

\section{Conclusion}
This study introduces a pioneering probabilistic forecasting approach utilizing CVAEs. The methodology centers on optimizing the CRPS as the loss function during training, allowing the CVAEs to learn predictive distributions effectively without imposing restrictive assumptions. Demonstrating both remarkable and consistent performance in our extensive empirical evaluations, this approach emerges as a significant and innovative contribution to the field of probabilistic forecasting, offering a robust tool for decision-makers to navigate the realm of uncertainty with confidence.
\bibliographystyle{unsrt}  
\bibliography{references}

\newpage
\appendixtitleon
\appendixtitletocon
\begin{appendices}
\section{Datasets}
\label{app:ds_rev}
\begin{figure*}
     \centering
     \begin{subfigure}{0.32\textwidth}
         \centering
         \includegraphics[width=\textwidth]{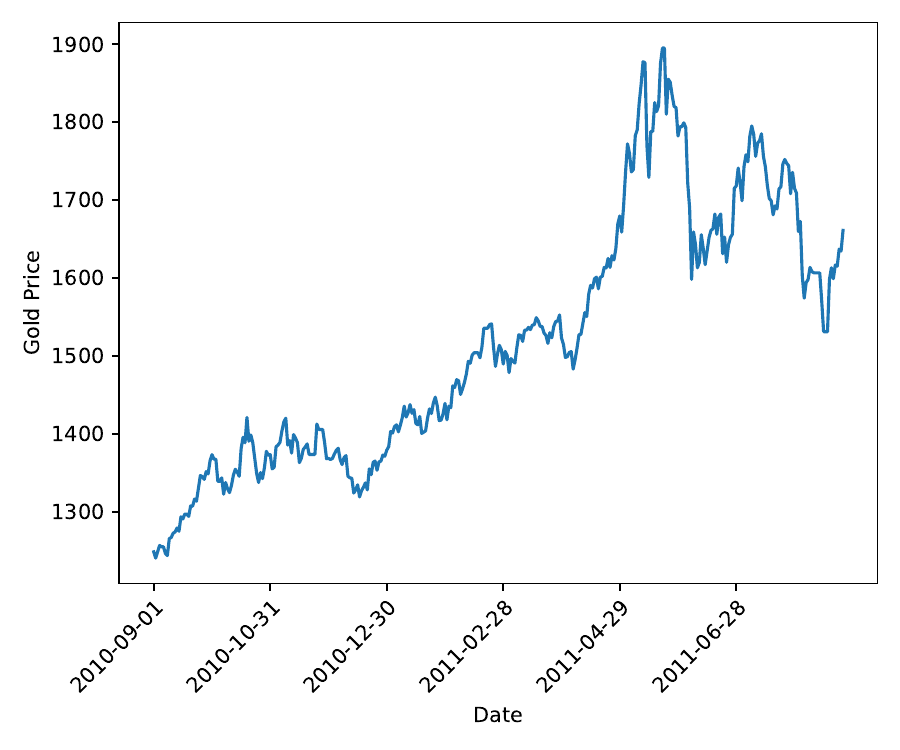}
         \caption{Gold Price Dataset}
     \end{subfigure}
     \hfill
     \begin{subfigure}{0.32\textwidth}
         \centering
         \includegraphics[width=\textwidth]{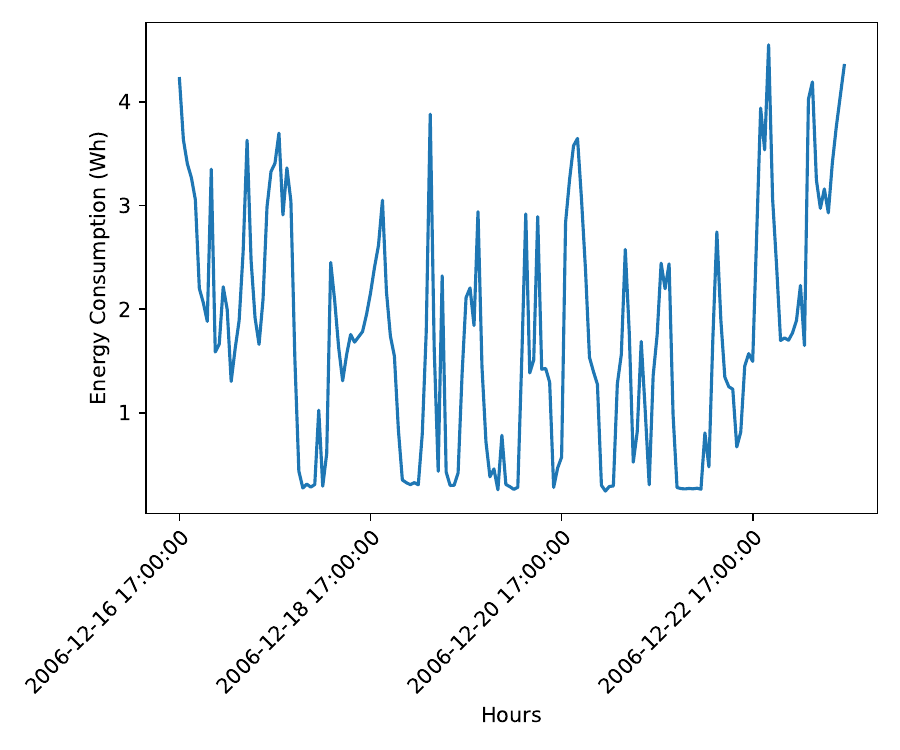}
         \caption{HEPC Dataset}
     \end{subfigure}
     \hfill
    \begin{subfigure}{0.32\textwidth}
         \centering
         \includegraphics[width=\textwidth]{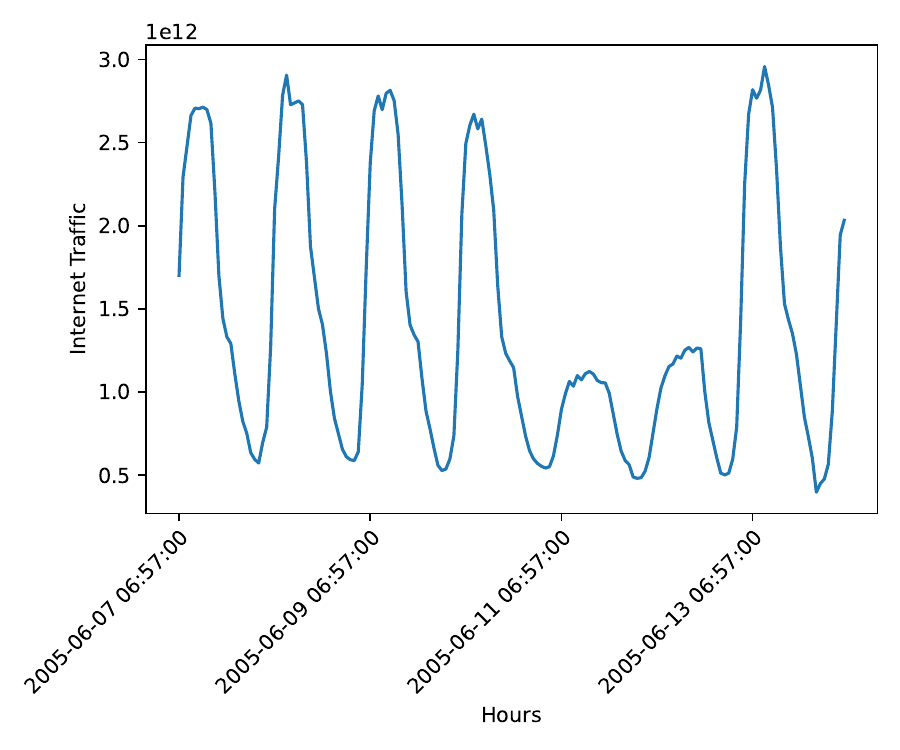}
         \caption{Internet Traffic A1H Dataset}
     \end{subfigure}
     \hfill
     \begin{subfigure}{0.32\textwidth}
         \centering
         \includegraphics[width=\textwidth]{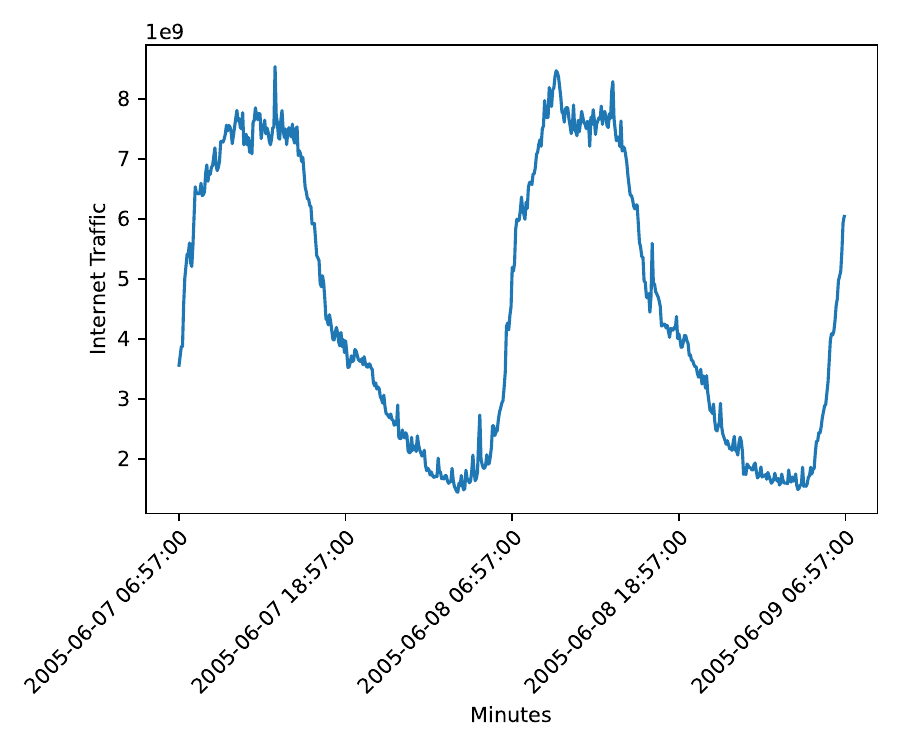}
         \caption{Internet Traffic A1M Dataset}
     \end{subfigure}
     \hfill
     \begin{subfigure}{0.32\textwidth}
         \centering
         \includegraphics[width=\textwidth]{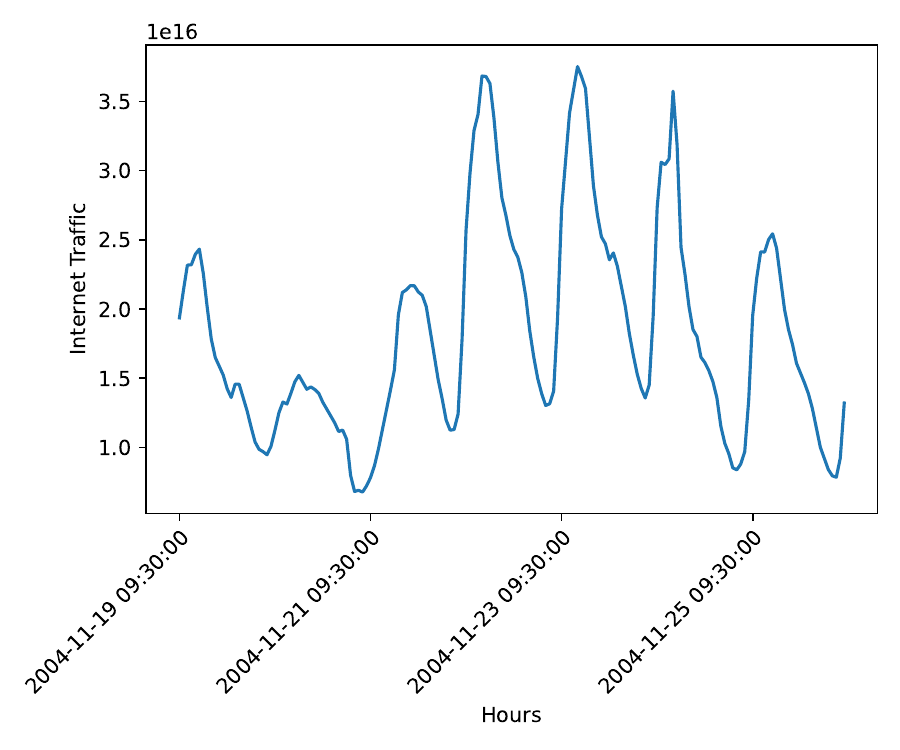}
         \caption{Internet Traffic B1H Dataset}
     \end{subfigure}
     \hfill
    \begin{subfigure}{0.32\textwidth}
         \centering
         \includegraphics[width=\textwidth]{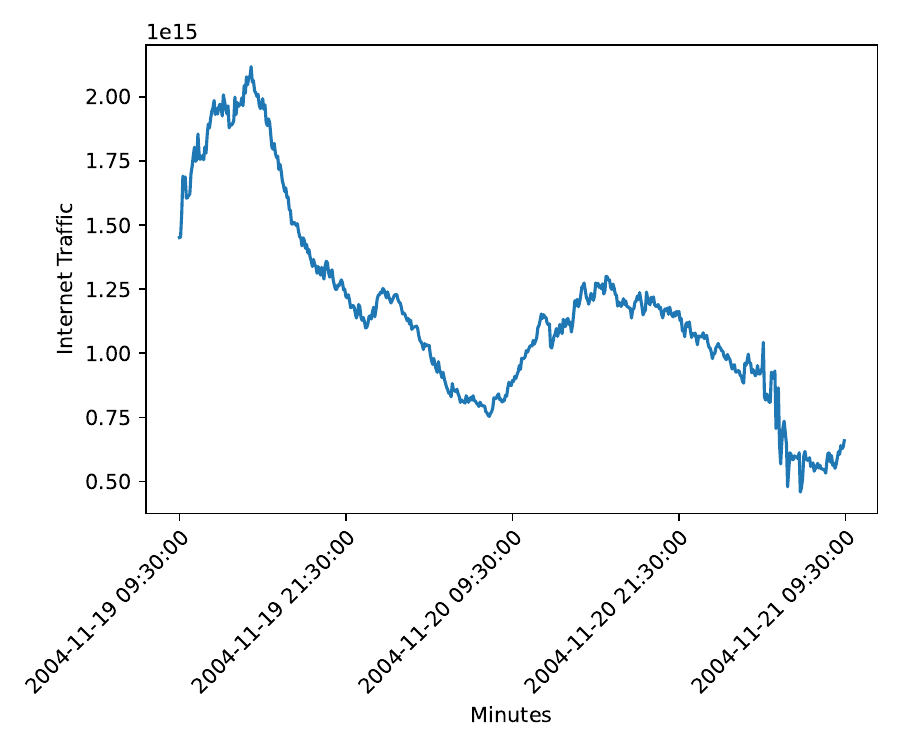}
         \caption{Internet Traffic B5M Dataset}
     \end{subfigure}
     \hfill
    \begin{subfigure}{0.32\textwidth}
         \centering
         \includegraphics[width=\textwidth]{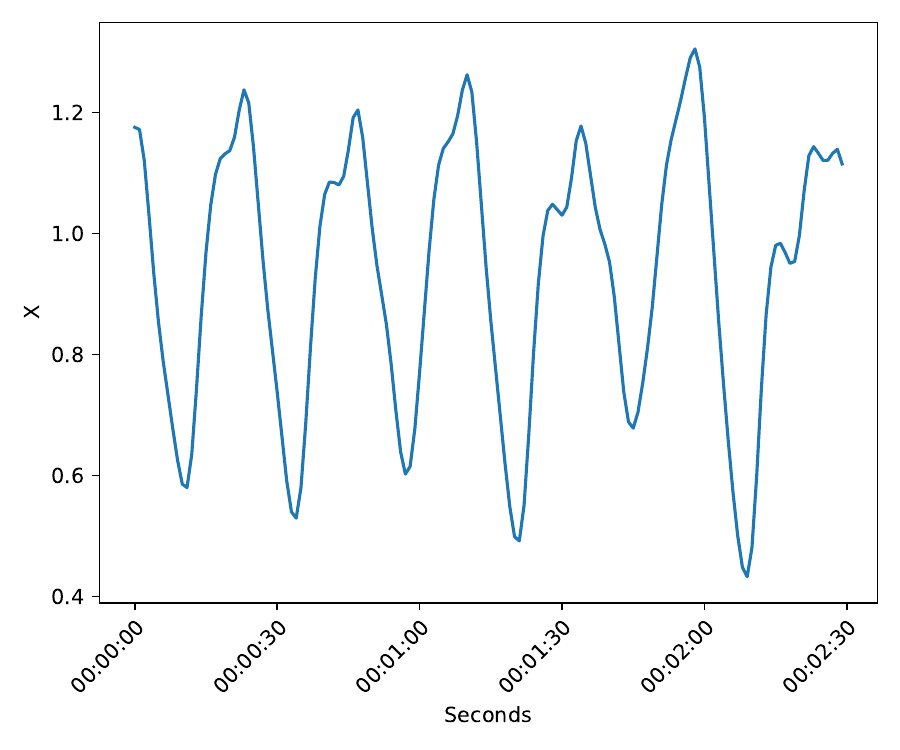}
         \caption{Mackey Glass Dataset}
     \end{subfigure}
     \hfill
     \begin{subfigure}{0.32\textwidth}
         \centering
         \includegraphics[width=\textwidth]{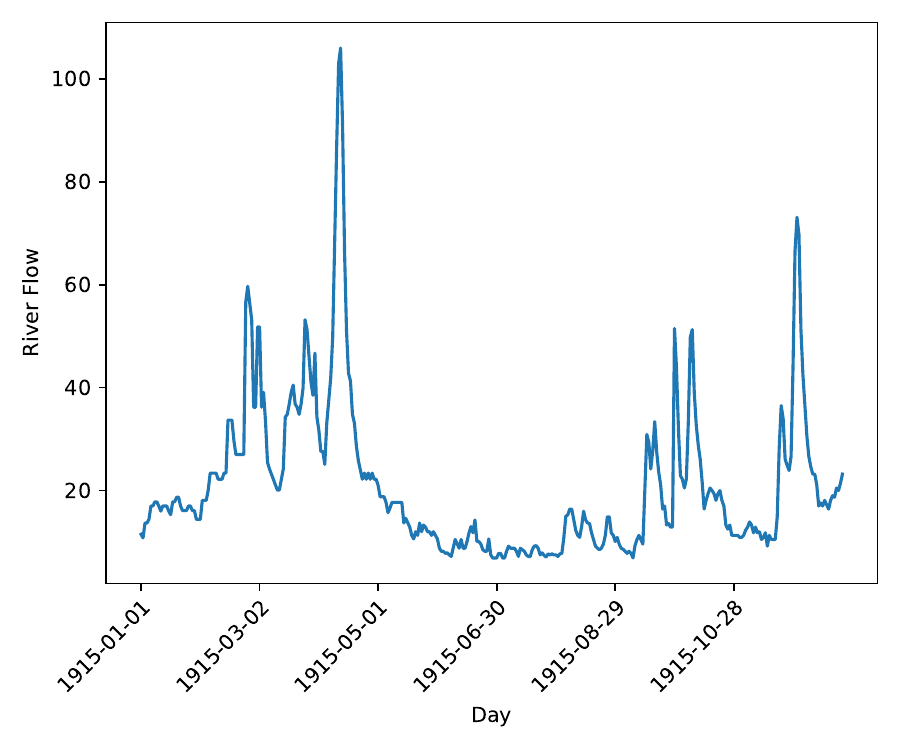}
         \caption{Saugeenday River Dataset}
     \end{subfigure}
     \hfill
    \begin{subfigure}{0.32\textwidth}
         \centering
         \includegraphics[width=\textwidth]{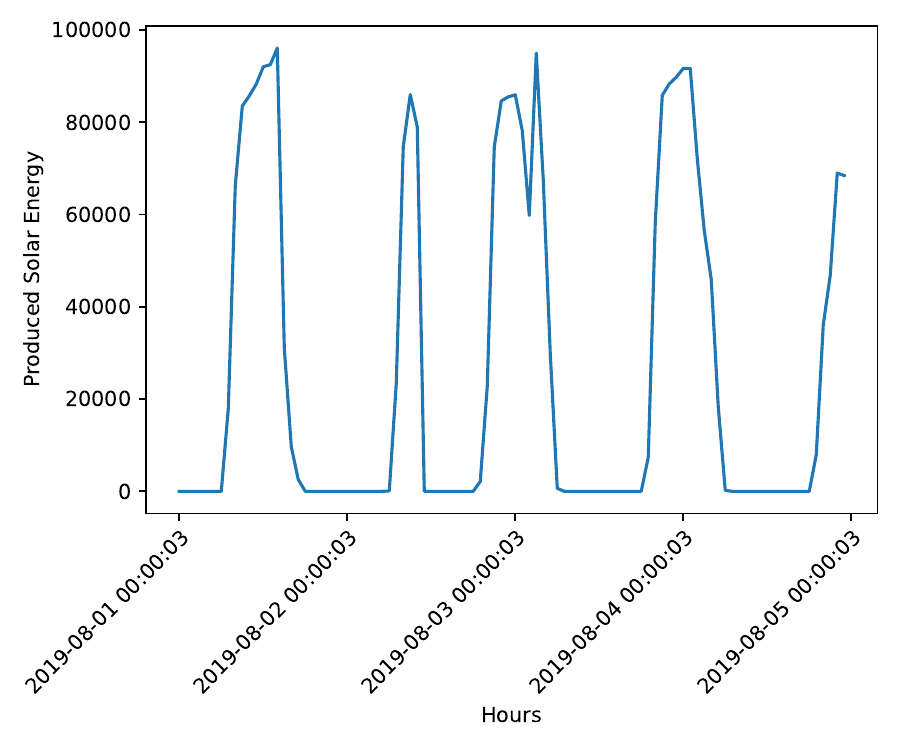}
         \caption{Solar-4-seconds Dataset}
     \end{subfigure}
          \hfill
    \begin{subfigure}{0.32\textwidth}
         \centering
         \includegraphics[width=\textwidth]{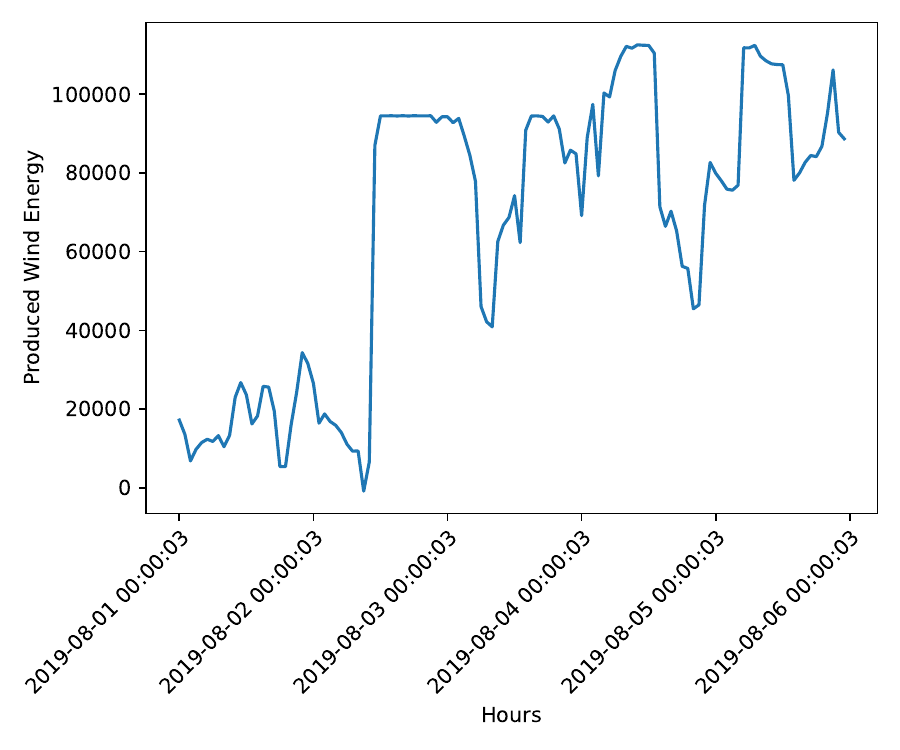}
         \caption{Wind-4-seconds Dataset}
     \end{subfigure}
     \hfill
     \begin{subfigure}{0.32\textwidth}
         \centering
         \includegraphics[width=\textwidth]{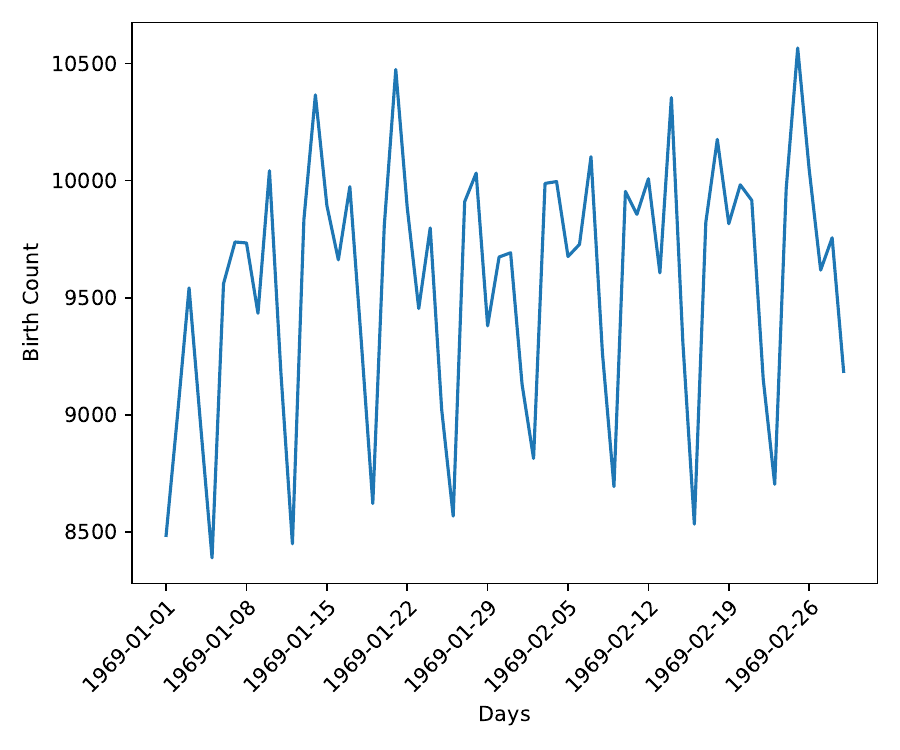}
         \caption{US Births Dataset}
     \end{subfigure}
     \hfill
    \begin{subfigure}{0.32\textwidth}
         \centering
         \includegraphics[width=\textwidth]{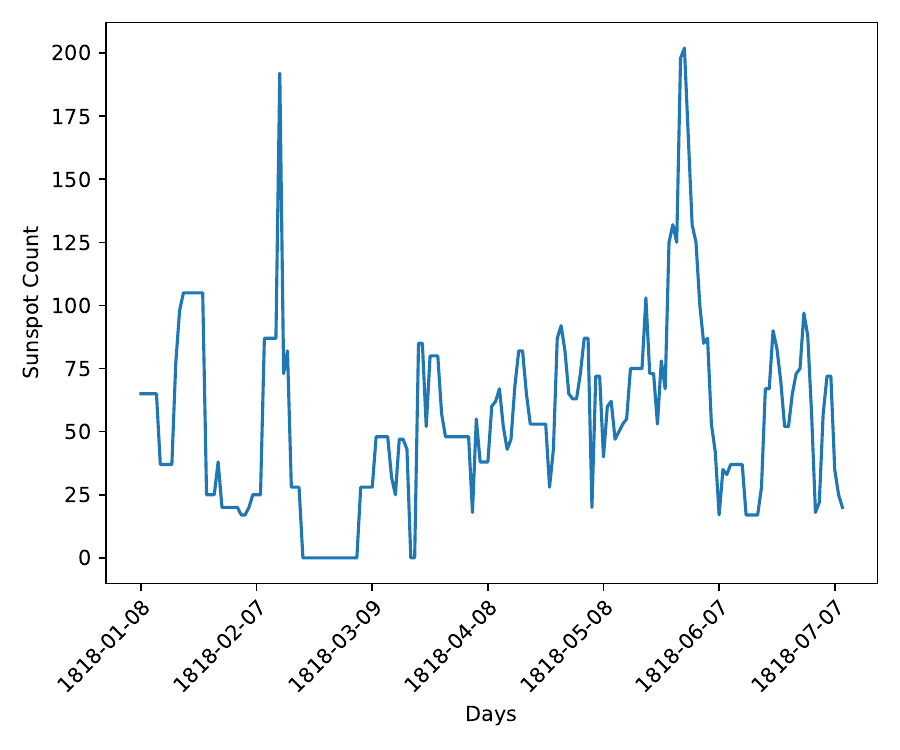}
         \caption{Sunspot Dataset}
     \end{subfigure}
        \caption{Illustration of a slice of each univariate time series dataset}
        \label{fig:app_dataset_slices}
\end{figure*}

This appendix provides further information on the datasets employed in this study. Table~\ref{tab:uni_prop} lists the main features of these datasets, and figure~\ref{fig:app_dataset_slices} visualizes a segment from each to provide an overview of the diverse dynamics of these datasets. The rest of this appendix presents further details for each dataset.

\begin{table}[h]
\vskip 0.15in
\begin{center}
\begin{scriptsize}
\begin{sc}
\caption{The properties of univariate datasets utilized for multi-step-ahead forecasting experiments}
\label{tab:uni_prop}
\begin{tabular}{@{}lllcc@{}}
\toprule
Dataset                         &  & & History  & \\
Name                             & Frequency & Length &  Size & Horizon\\\midrule
Gold Price             & Daily     & 2487   & 60                & 30\\
HPEC                 & Hourly    & 34569  & 48                & 24 \\
Internet Traffic A1H & Hourly    & 1231   & 48                & 24 \\
Internet Traffic A5M & 5 minutes & 14772  & 24                & 12 \\
Internet Traffic B1H & Hourly    & 1657   & 48                & 24 \\
Internet Traffic B5M & 5 minutes & 19888  & 24                & 12 \\
Mackey Glass         & Seconds   & 20000  & 120               & 60 \\
Saugeen River Flow   & Daily     & 23741  & 60                & 30 \\
Sunspot              & Daily     & 73924  & 60                & 30 \\
US Births            & Daily     & 7305   & 60                & 30 \\
Solar                & Hourly    & 8219   & 48                & 24 \\
Wind                 & Hourly    & 8219   & 48                & 24 \\ \bottomrule
\end{tabular}
\end{sc}
\end{scriptsize}
\end{center}
\vskip -0.1in
\end{table}

\subsection{Gold Price Dataset}
\label{sec:gold_dataset}
Gold is a vital commodity with a significant history of global trading. The Gold Price dataset\footnote{Accessible in \url{www.kaggle.com/datasets/arashnic/learn-time series-forecasting-from-gold-price}} provides daily pricing details, encompassing 2,487 data points from September 1st, 2010, to March 13th, 2020. This dataset facilitates analytical and predictive modeling of the gold market due to its comprehensive coverage during the specified period.

\subsection{Household Electric Power Consumption Dataset}
The Household Electric Power Consumption (HEPC) dataset~\cite{hebrail_individual_2012} comprises electric power consumption readings from a singular household with a sampling rate of one minute. From December 16th, 2006, to November 26th, 2010, it encapsulates nearly four years of data. Notably, there are timestamps with missing values. An imputation approach, where missing values are replaced with the average of their adjacent observations, is adopted to ensure data integrity. Moreover, the dataset is resampled to aggregate the readings into hourly intervals, ensuring manageability while preserving inherent patterns.

\subsection{Internet Traffic Datasets}
Internet traffic datasets~\cite{cortez_multi-scale_2012} offer data from two distinct ISPs, named A and B. The A dataset pertains to a private ISP with nodes across 11 European cities, capturing data on a transatlantic link from June 7the to July 29the, 2005. The B dataset is sourced from UKERNA1 and aggregates traffic across the UK's academic network from November 19th, 2004, to January 27th, 2005. The A dataset logs every 30 seconds, whereas B records every 5 minutes. Aggregated variants, A5M, A1H, B5M, and B1H, offer data at 5-minute and 1-hour resolutions respectively.

\subsection{Macky-Glass Dataset}
The time-delay differential equation proposed by Mackey and Glass~\cite{mackey_oscillation_1977}\footnote{The dataset can be accessed at \url{https://git.opendfki.de/koochali/forgan}} stands as a benchmark for generating chaotic time series. The equation is given by:
\begin{equation}
\label{eq:mg}
\dot{x} = \frac{a\,x(t - \tau)}{(1 + 10\cdot(t - \tau)) - b\,x(t)} \,.
\end{equation}

Adhering to the parameters in~\cite{mendez_competitive_2017}, where $a = 0.1$, $b = 0.2$, and $\tau = 17$, a dataset of length 20,000 is generated using Eq.~(\ref{eq:mg}.

\subsection{Saugeen River Flow Dataset}
This dataset~\cite{ogie_optimal_2017} includes a univariate time series showcasing the daily mean flow of the Saugeen River at Walkerton measured in cubic meters per second (\(\frac{m^3}{s}\)). It captures data from January 1st, 1915, to December 31st, 1979. The dataset is archived in the Monash Time Series Forecasting Archive~\cite{godahewa_monash_2021}.

\subsection{Solar-4-seconds and Wind-4-seconds datasets}
These datasets capture solar and wind power production every 4 seconds from August 1st, 2019, for approximately one year. For the purpose of this research, the datasets are aggregated to represent hourly resolutions. Sourced from the Australian Energy Market Operator (AEMO) online platform, they are also part of the Monash Time Series Forecasting Archive~\cite{godahewa_monash_2021}.

\subsection{Sunspot Dataset}
The dataset contains a time series depicting daily sunspot counts from January 8th, 1818, to May 31st, 2020. Any missing values within the dataset are addressed using the Last Observation Carried Forward (LOCF) method. The Solar Influences Data Analysis Center is the primary data source, with the dataset also being part of the Monash Time Series Forecasting Archive~\cite{godahewa_monash_2021}.

\subsection{US Birth Dataset}
This dataset illustrates the number of births in the US from January 1st, 1969, to December 31st, 1988. Originally sourced from the R mosaicData package~\cite{pruim_mosaicdata_2022}, it's accessed for this research via the Monash Time Series Forecasting Archive~\cite{godahewa_monash_2021}.

\section{Baselines}
\label{app:baseline}
This section is dedicated to a detailed examination of 12 advanced probabilistic forecasting models. These models represent the latest advancements in the field and have demonstrated their effectiveness and robustness across various research studies. The primary focus will be on elucidating the distinctive features, methodologies, and underlying principles of each model, highlighting their contributions to advancing probabilistic forecasting. In subsequent chapters, this paper will benchmark these models as baselines, employing them as comparative standards to evaluate the performance of newly proposed methodologies in diverse experimental settings. This comparative analysis aims to provide a comprehensive understanding of where the newly proposed models stand in the context of current state-of-the-art probabilistic forecasting techniques.

\subsection{DeepAR}
\label{sec:deepar}

DeepAR, as presented by Salinas et al.~\cite{salinas_deepar_2020}, serves as a prominent explicit probabilistic forecasting model in the realm of time series prediction. At its core, DeepAR constructs a predictive distribution utilizing a Gaussian distribution, wherein the distribution's parameters, specifically the mean and standard deviation, are derived from the historical data.

To facilitate this parameter extraction, DeepAR integrates a neural network structure. This architecture predominantly features a multi-layer Recurrent Neural Network (RNN) to map the time series history onto the Gaussian distribution parameters. The training regimen for the network adopts a maximum likelihood estimation paradigm. Within this paradigm, the probability density function of the Gaussian distribution acts as the likelihood function, ensuring optimal parameter approximation.

\subsection{DeepState}
Rangapuram et al.~\cite{rangapuram_deep_2018} proposed the DeepState model as a novel approach that combines state space models (SSMs) with deep learning for probabilistic time series forecasting. The paper proposes bridging the gap between SSMs and deep learning by parametrizing a linear SSM with a jointly-learned recurrent neural network (RNN). The deep recurrent neural network parametrizes the mapping from covariates to state space model parameters. The model is interpretable, as it allows inspection and modification of SSM parameters for each time series. It can automatically extract features and learn complex temporal patterns from raw time series and associated covariates. The model promises to contain the interpretability and data efficiency of SSMs and large data handling of deep neural networks.

\subsection{DeepFactor}
The DeepFactor~\cite{wang_deep_2019} employs both classical and deep neural time series forecasting models to construct probabilistic forecaster. DeepFactor is a global-local method where each time series or its latent function is represented as a combination of a global time series and a corresponding local model. The global factors are modeled by Recurrent Neural Networks (RNNs). These latent global deep factors can be thought of as dynamic principal components driving the underlying dynamics of all time series. Local models can include various choices like white noise processes, linear dynamical systems (LDS), or Gaussian processes (GPs). This stochastic component captures individual random effects for each time series. DeepFactor systematically combines of deep neural networks and probabilistic models in a global-local framework and develops an efficient and scalable inference algorithm for non-Gaussian likelihoods.

\subsection{Deep Renewal Processes}
Deep Renewal Processes~\cite{turkmen_forecasting_2021} is a novel framework for probabilistic forecasting of intermittent and sparse time series. The method focuses on forecasting demand data that appears sporadically, posing unique challenges due to long periods of zero demand followed by non-zero demand. The framework is based on discrete-time renewal processes, which are adept at handling patterns like aging, clustering, and quasi-periodicity in demand arrivals. The model innovatively incorporates neural networks, specifically recurrent neural networks (RNNs), replacing exponential smoothing with a more flexible neural approach. Moreover, the framework also extends to model continuous-time demand arrivals, enhancing flexibility and applicability.

\subsection{GPForecaster}
For the Gaussian Process (GP), we employed a model implemented in GluonTS~\cite{alexandrov_gluonts_2020}. This model defines a Gaussian Process with Radial Basis Function (RBF) as the kernel for each time step. GP with an RBF kernel can capture both the trends and nuances of time series data, while its probabilistic nature provides a measure of uncertainty or confidence in its predictions. This combination is especially valuable in scenarios where understanding the uncertainty of future events is as crucial as the predictions themselves.

\subsection{Multi-Horizon Quantile Recurrent Forecaster (MQ-RNN)}
The Multi-Horizon Quantile Recurrent Forecaster, or in short MQ-RNN~\cite{wen_multi-horizon_2018}, is a direct quantile regression method that aims to forecast the quantile of predictive distribution for multiple horizons directly. The model employs Seq2Seq architecture for quantile regression. Besides MQ-RNN, this paper also utilized MQ-CNN, which is the same model with CNN components as the main components of Seq2Seq architecture instead of RNN modules.

\subsection{Multi-Horizon Quantile Convolutional Forecaster (MQ-CNN)}
MQ-CNN is a similar method to MQ-RNN, with CNN components as the main components of Seq2Seq architecture instead of RNN modules.

\subsection{Prophet}
The prophet~\cite{taylor_forecasting_2018} is a robust and scalable solution for forecasting time series data, particularly in business contexts where handling specific scenarios such as multiple strong seasonalities, trend changes, outliers, and holiday effects becomes paramount. The prophet decomposes time series into three components. The first component, trend, Captures non-periodic changes in the data. The second component, seasonality, represents periodic changes, like weekly yearly cycles. finally, the last component, holidays, accounts for holidays and events that occur on irregular schedules. Each component is modeled separately and fits using Bayesian infeProbabilistic Stream Flow
Forecasting for Reservoir Operatiorence to estimate parameters. Furthermore, the model is designed to be intuitive, allowing analysts to make informed adjustments based on domain knowledge.

\subsection{WaveNet}
\label{sec:wavenet}
The Wavenet~\cite{oord_wavenet_2016} is a powerful generative model originally proposed for voice synthesis. Wavenet forecasts sequentially in an autoregressive fashion using previously generated samples as input to predict the next.WaveNet employs a deep neural network consisting of stacks of dilated causal convolutions, enabling it to capture a wide range of audio frequencies and temporal dependencies. Wavenet defines predictive distribution for each time step in the horizon by quantifying the time series and using a softmax at the output layer of the model. In other words, the wavenet quantifies the time series to make the problem similar to a classification problem and provides the output of the softmax layer as the predictive distribution.

\subsection{Transformer}
The transformer~\cite{vaswani_attention_2017} is a novel sequence-to-sequence model that represents a departure from previous methods by relying entirely on attention mechanisms, dispensing with recurrent and convolutional neural networks. The Transformer follows the typical encoder-decoder structure, but both the encoder and the decoder are composed of a stack of identical layers, each with a novel self-attention mechanism. The Self-Attention mechanism allows the model to weigh the importance of different time steps in the input sequence, enabling it to capture context more effectively. At the core of the model is the scaled dot-product attention, a mechanism that calculates the attention scores based on the scaled dot products of queries and keys. Unlike recurrent models, the Transformer allows for much more parallelization, making training faster. Moreover, the self-attention mechanism enables the model to learn long-range dependencies more effectively.

\subsection{Temporal Fusion Transformers}
Temporal Fusion Transformers (TFT)~\cite{lim_temporal_2021} is an encoder-decoder-based quantile regression model for multi-horizon forecasting. It has many novelties under the hood that make its performance excel in many scenarios, including:
\begin{itemize}
    \item \textbf{Static Covariate Encoders}: These encode static context vectors, which are crucial for conditioning the temporal dynamics within the network.
    \item \textbf{Gating Mechanisms}: They provide adaptive depth and complexity to the network, allowing it to handle a wide range of datasets and scenarios.
    \item \textbf{Variable Selection Networks}: These networks select relevant input variables at each time step, enhancing interpretability and model performance by focusing on the most salient features.
    \item \textbf{Temporal Processing}: The model employs a sequence-to-sequence layer for local processing and interpretable multi-head attention mechanisms to capture long-term dependencies.
\end{itemize}

The TFT demonstrates superior performance across various real-world datasets, outperforming existing benchmarks in multi-horizon forecasting. The model facilitates interpretability, allowing users to identify globally important variables, persistent temporal patterns, and significant events.

\subsection{ForGAN}
ForGAN~\cite{koochali_probabilistic_2019} is a probabilistic forecaster that employs a conditional GAN to learn a mapping from a prior distribution to the predictive distribution conditioned on an input history window. It incorporates Recurrent Neural Network (RNN) modules in both the Generator and Discriminator components. The ForGAN was proposed for one-step ahead univariate time series forecasting. In this paper, we extend the model to encompass multi-step ahead forecasting by incorporating an auto-regressive approach.

\section{VAEneu Hyperparameters}
\label{app:hyp}
The VAEneu hyperparameters are either fixed values or defined solely based on history windows size (HWS), i.e., the condition. Table~\ref{tab:hyp} lists the VAEneu models' hyperparameters.

\begin{table}
\vskip 0.15in
\begin{center}
\begin{scriptsize}
\begin{sc}
\caption{Hyperparameters of VAEneu models}
\label{tab:hyp}
\begin{tabular}{ll}
\toprule
Hyperparameter   & Value                \\ \midrule
\multicolumn{2}{l}{VAEneu-RNN}           \\\midrule
Sample size   &  8   \\
Number of LSTM layers & 1                    \\
LSTM hidden size      & $\frac{HWS}{2}$      \\
\addlinespace[0.5em]
Latent variable~"\(z\)" size       & $\frac{HWS}{2}$      \\

\midrule

\multicolumn{2}{l}{VAEneu-TCN}           \\\midrule
Sample size   &  8   \\
TCN block kernel size      & 5                    \\

TCN block number of layers & Log($\frac{HWS}{8}$) \\
\addlinespace[0.5em]
TCN block hidden size      & 2$\times$Log(HWS)    \\

Latent variable~"\(z\)" size       & 2$\times$Log(HWS)    \\ \bottomrule
\end{tabular}
\end{sc}
\end{scriptsize}
\end{center}
\vskip -0.1in
\end{table}

\section{Critical Difference}
\label{app:cd}
For the computation of the critical difference depicted in our diagram, we rely on the outcomes from three statistical tests. This study restricts statistical significance at the 0.05 threshold level ($\alpha = 0.05$).

\paragraph{Friedman test:}
The initial analytical step entails the application of the Friedman test, aimed at discerning any significant differences among the algorithms spanning multiple datasets. The null hypothesis, denoted as $\pmb{H_0}$, posits the equivalence in performance across all tested models. Contrarily, the alternative hypothesis, $\pmb{H_1}$, suggests a differential performance exhibited by at least one algorithm relative to the others. The rejection of $\pmb{H_0}$ in favor of $\pmb{H_1}$ signifies the potential utility of the CD diagram, thereby certifying the progression to subsequent statistical tests.

\paragraph{Wilcoxon signed-rank test:}
For pairwise comparisons of model performances across all datasets, the study employs the Wilcoxon signed-rank test. This non-parametric test facilitates the comparison of two paired samples without assuming that the differences between said pairs adhere to a normal distribution. Given two models, A and B, with their performances represented as sets of $\overline{CRPS}$ values spanning the datasets, the null hypothesis, $\pmb{H_0}$, asserts a median difference of zero between paired $\overline{CRPS}$ values, thereby suggesting similar performance between Models A and B across all dataset. The outcome, be it acceptance or rejection of $\pmb{H_0}$, is utilized for defining the critical difference band within the CD diagram.

\paragraph{Paired t-test:}
Finally, the paired t-test is applied to rank model performances within specific datasets. A paired t-test is used to compare the means of two related groups when the samples are dependent; that is, each observation in one sample can be paired with an observation in the other sample. Within the scope of model performance evaluated over multiple runs, these dependent observations represent the performances of two distinct models assessed on an identical dataset across separate runs. Let $\overline{CRPS}_A$ symbolize the mean CRPS of model A, computed over three runs on a specific dataset. If the inequality $\overline{CRPS}_A~<~\overline{CRPS}_B$ holds for a dataset, then the null hypothesis $\pmb{H_0}$ of the one-sided paired t-test posits a superior or equivalent performance by Model B relative to Model A across all runs. The alternative hypothesis, $\pmb{H_1}$, counters this by indicating a statistically superior performance by Model A. We sequentially applied this test to every possible model pair within a dataset to determine their ranks. In scenarios where the null hypothesis faced rejection, Model A was accorded a superior rank. Conversely, an acceptance led to the assignment of identical ranks to both Model A and Model B. Subsequent to the ranking procedure, the models' average ranks across all datasets were computed to establish their respective positions on the CD diagram.

\section{Qualitative Result}
\label{app:qr}
\begin{figure*}
     \captionsetup[subfigure]{labelformat=empty}
     \centering
    \begin{subfigure}{0.32\textwidth}
         \centering
         \includegraphics[width=\textwidth]{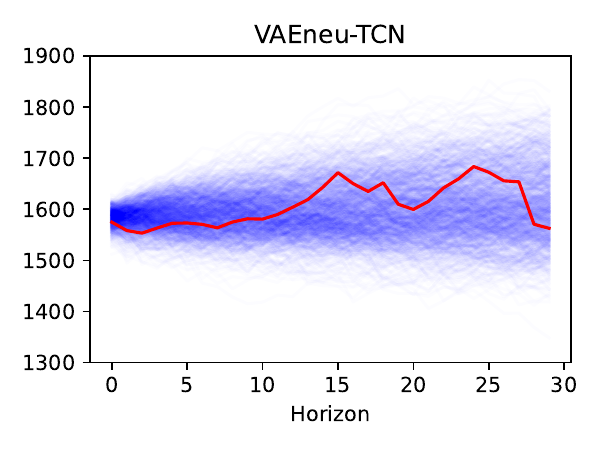}
         \caption{~}
     \end{subfigure}
     \hfill
     \begin{subfigure}{0.32\textwidth}
         \centering
         \includegraphics[width=\textwidth]{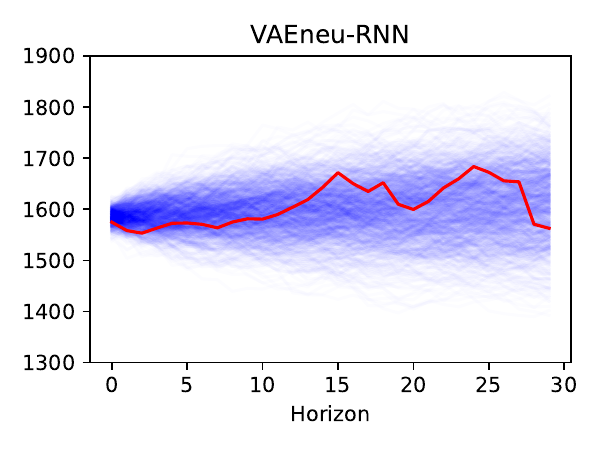}
         \caption{Gold Price Dataset}
     \end{subfigure}
     \hfill
    \begin{subfigure}{0.32\textwidth}
         \centering
         
         \includegraphics[width=\textwidth]{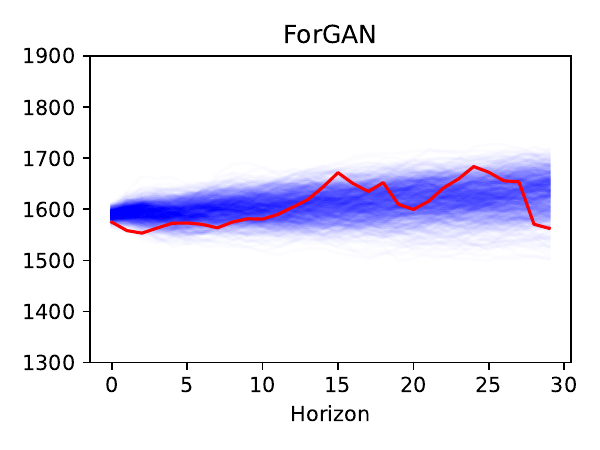}
         \caption{~}
     \end{subfigure}
     
     \vfill
    
    \begin{subfigure}{0.32\textwidth}
         \centering
         \includegraphics[width=\textwidth]{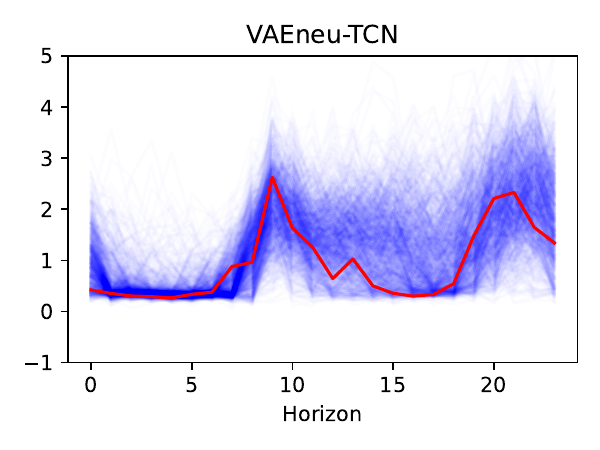}
         \caption{~}
     \end{subfigure}
     \hfill
     \begin{subfigure}{0.32\textwidth}
         \centering
         \includegraphics[width=\textwidth]{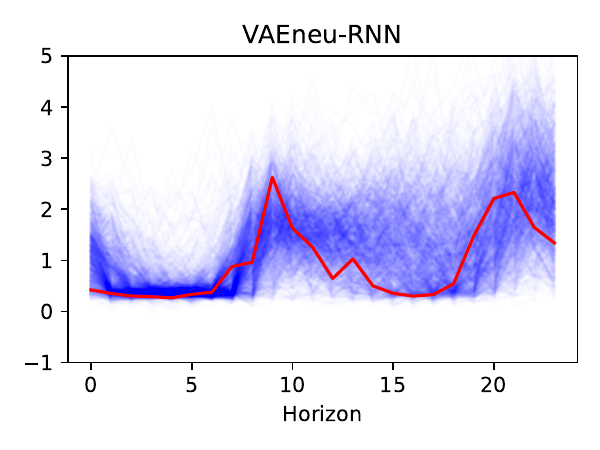}
         \caption{HEPC Dataset}
     \end{subfigure}
     \hfill
    \begin{subfigure}{0.32\textwidth}
         \centering
         
         \includegraphics[width=\textwidth]{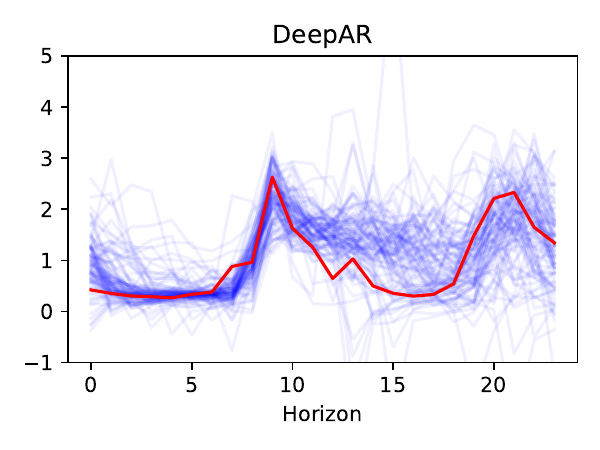}
         \caption{~}
     \end{subfigure}

     \vfill
    
    \begin{subfigure}{0.32\textwidth}
         \centering
         \includegraphics[width=\textwidth]{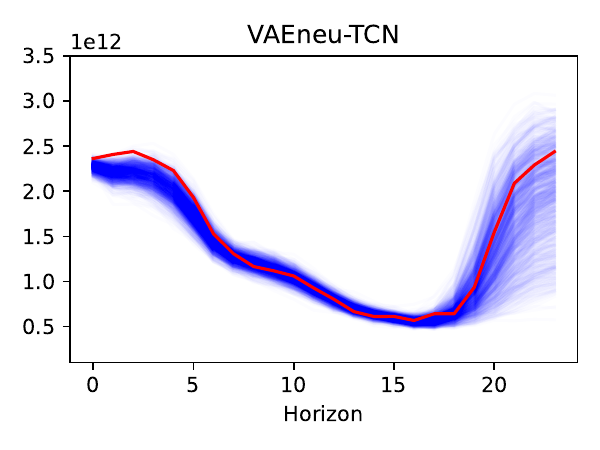}
         \caption{~}
     \end{subfigure}
     \hfill
     \begin{subfigure}{0.32\textwidth}
         \centering
         \includegraphics[width=\textwidth]{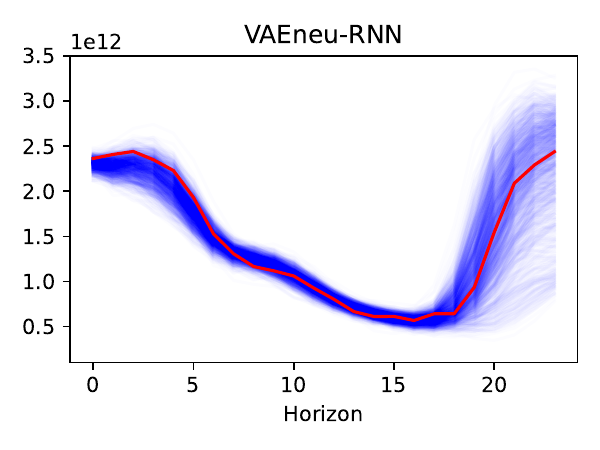}
         \caption{Internet Traffic A1H Dataset}
     \end{subfigure}
     \hfill
    \begin{subfigure}{0.32\textwidth}
         \centering
         
         \includegraphics[width=\textwidth]{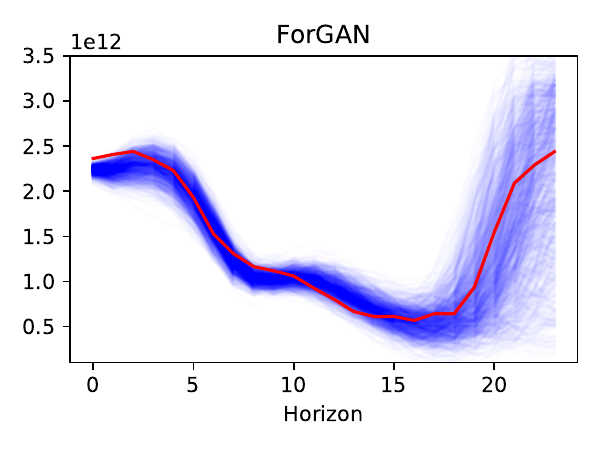}
         \caption{~}
     \end{subfigure}

     \vfill
    
    \begin{subfigure}{0.32\textwidth}
         \centering
         \includegraphics[width=\textwidth]{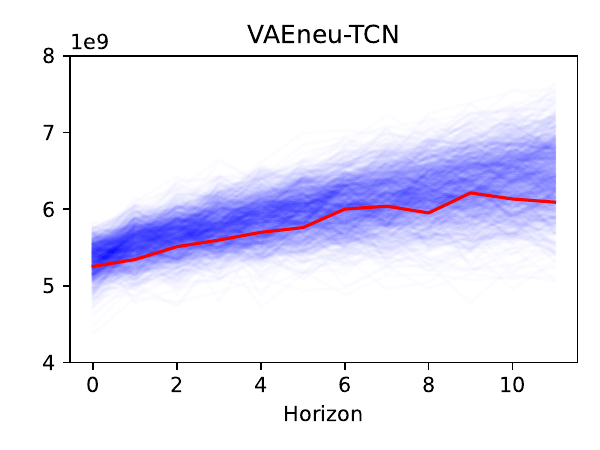}
         \caption{~}
     \end{subfigure}
     \hfill
     \begin{subfigure}{0.32\textwidth}
         \centering
         \includegraphics[width=\textwidth]{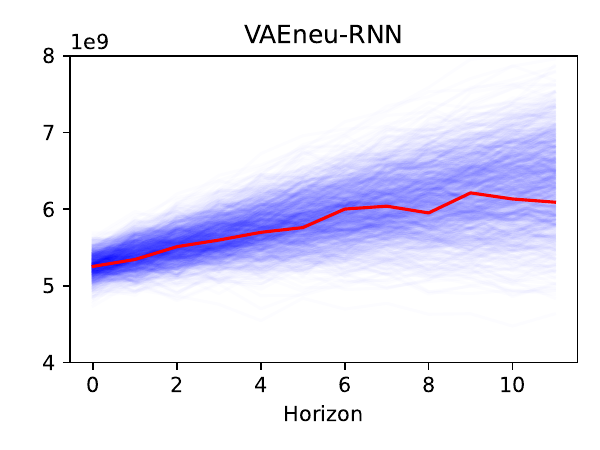}
         \caption{Internet Traffic A5M Dataset}
     \end{subfigure}
     \hfill
    \begin{subfigure}{0.32\textwidth}
         \centering
         
         \includegraphics[width=\textwidth]{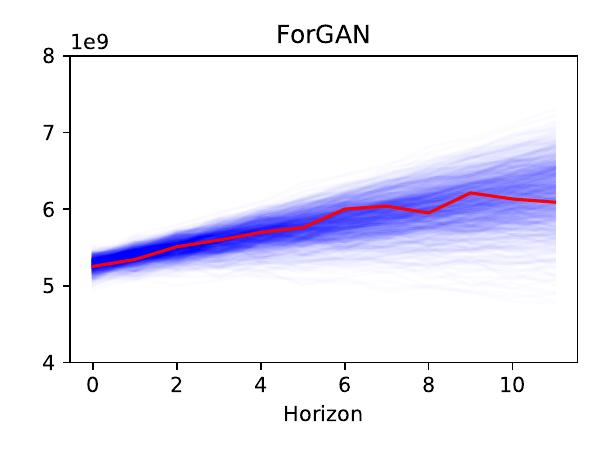}
         \caption{~}
     \end{subfigure}
        \caption{Sample of the forecast for VAEneu models alongside the best baseline model.}
        \label{fig:sample_appendix1}
\end{figure*}

\begin{figure*}
     \captionsetup[subfigure]{labelformat=empty}
     \centering
         \begin{subfigure}{0.32\textwidth}
         \centering
         \includegraphics[width=\textwidth]{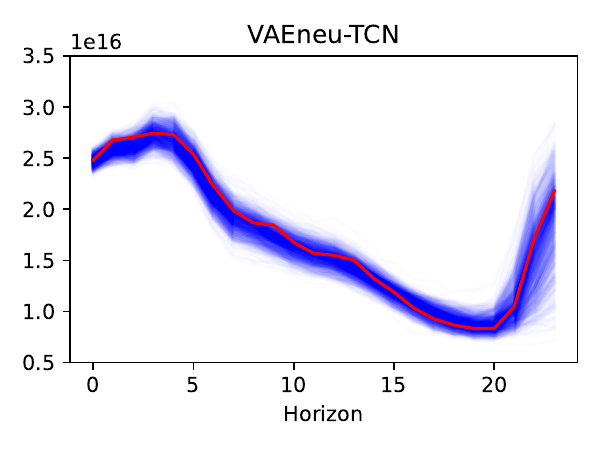}
         \caption{~}
     \end{subfigure}
     \hfill
     \begin{subfigure}{0.32\textwidth}
         \centering
         \includegraphics[width=\textwidth]{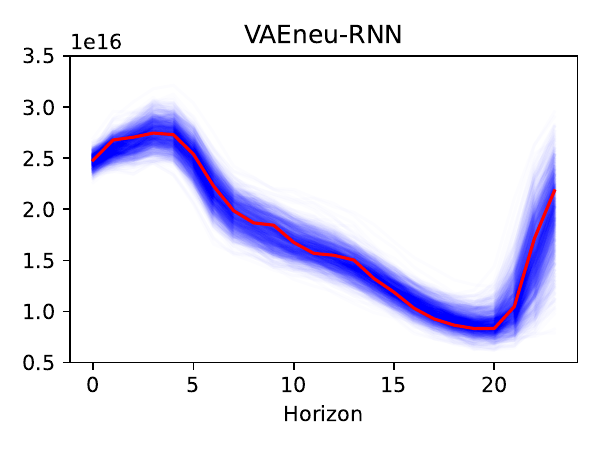}
         \caption{Internet Traffic B1H Dataset}
     \end{subfigure}
     \hfill
    \begin{subfigure}{0.32\textwidth}
         \centering
         
         \includegraphics[width=\textwidth]{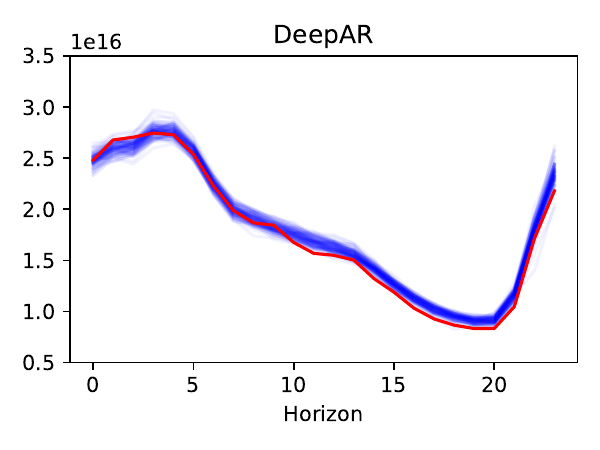}
         \caption{~}
     \end{subfigure}
     
     \vfill
    \begin{subfigure}{0.32\textwidth}
         \centering
         \includegraphics[width=\textwidth]{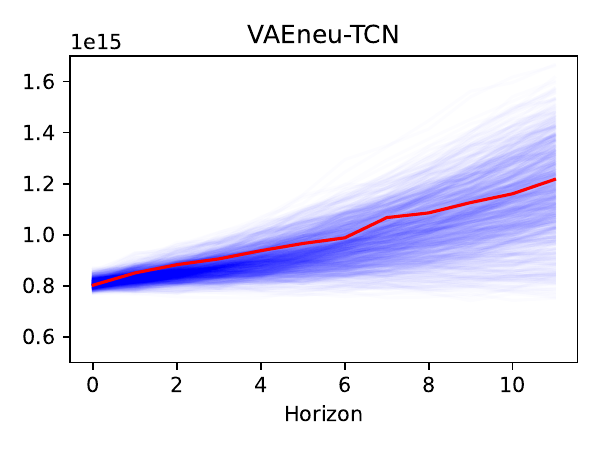}
         \caption{~}
     \end{subfigure}
     \hfill
     \begin{subfigure}{0.32\textwidth}
         \centering
         \includegraphics[width=\textwidth]{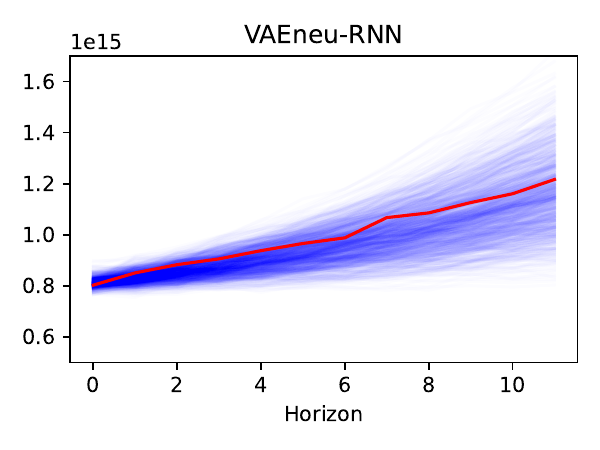}
         \caption{Internet Traffic B5M Dataset}
     \end{subfigure}
     \hfill
    \begin{subfigure}{0.32\textwidth}
         \centering
         
         \includegraphics[width=\textwidth]{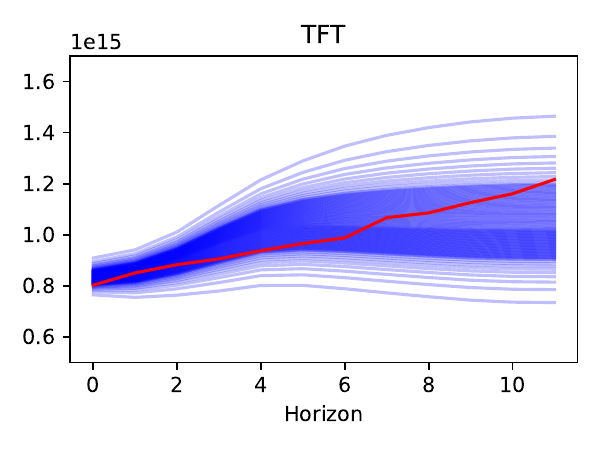}
         \caption{~}
     \end{subfigure}
     
     \vfill
    
    \begin{subfigure}{0.32\textwidth}
         \centering
         \includegraphics[width=\textwidth]{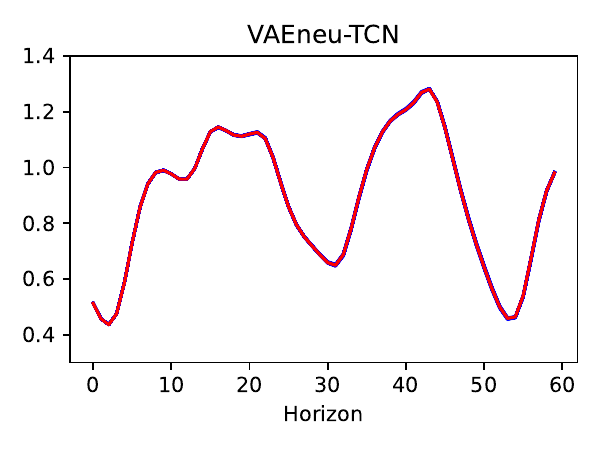}
         \caption{~}
     \end{subfigure}
     \hfill
     \begin{subfigure}{0.32\textwidth}
         \centering
         \includegraphics[width=\textwidth]{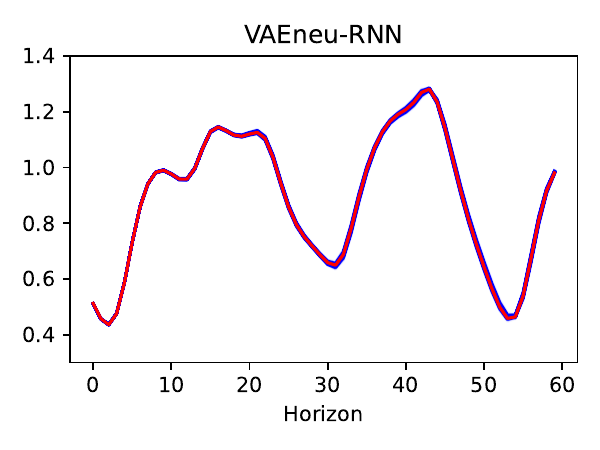}
         \caption{Mackey Glass Dataset}
     \end{subfigure}
     \hfill
    \begin{subfigure}{0.32\textwidth}
         \centering
         
         \includegraphics[width=\textwidth]{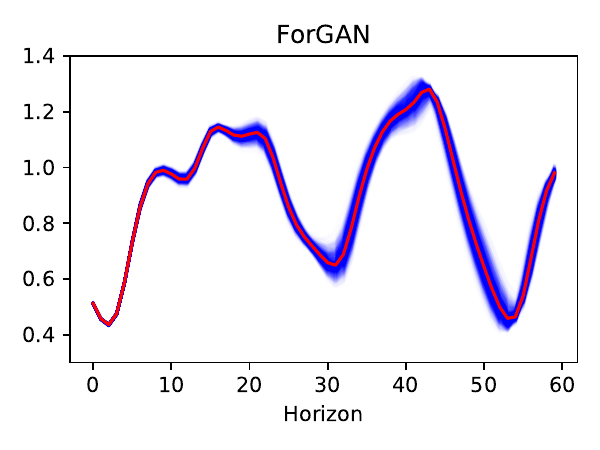}
         \caption{~}
     \end{subfigure}
     
     \vfill
    
    \begin{subfigure}{0.32\textwidth}
         \centering
         \includegraphics[width=\textwidth]{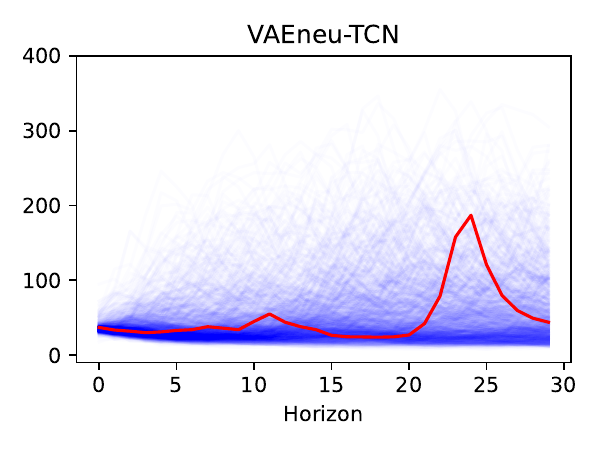}
         \caption{~}
     \end{subfigure}
     \hfill
     \begin{subfigure}{0.32\textwidth}
         \centering
         \includegraphics[width=\textwidth]{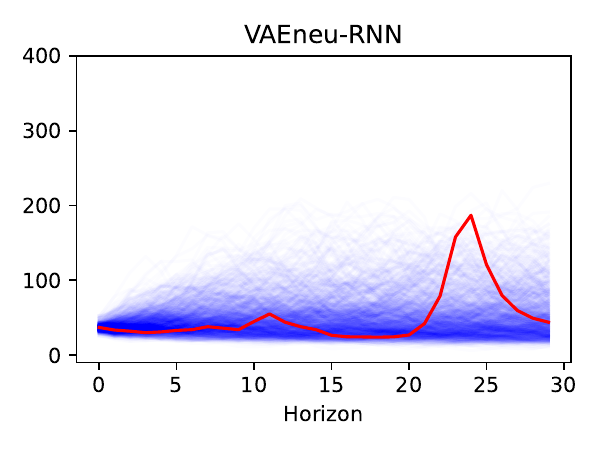}
         \caption{Saugeenday River Dataset}
     \end{subfigure}
     \hfill
    \begin{subfigure}{0.32\textwidth}
         \centering   
         \includegraphics[width=\textwidth]{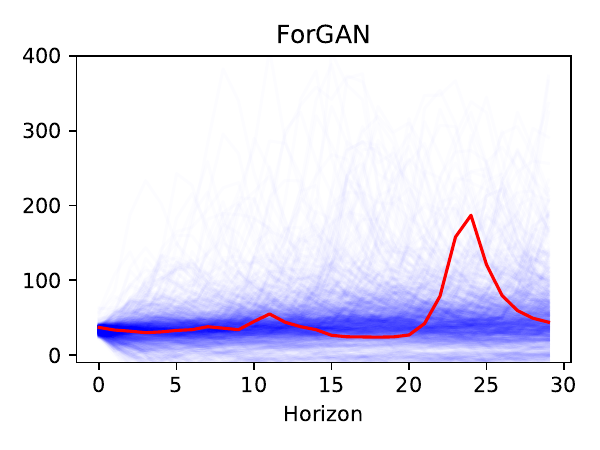}
         \caption{~}
     \end{subfigure}
        \caption{Sample of the forecast for VAEneu models alongside the best baseline model.}
        \label{fig:sample_appendix2}
\end{figure*}

\begin{figure*}
     \captionsetup[subfigure]{labelformat=empty}
     \centering
    \begin{subfigure}{0.32\textwidth}
         \centering
         \includegraphics[width=\textwidth]{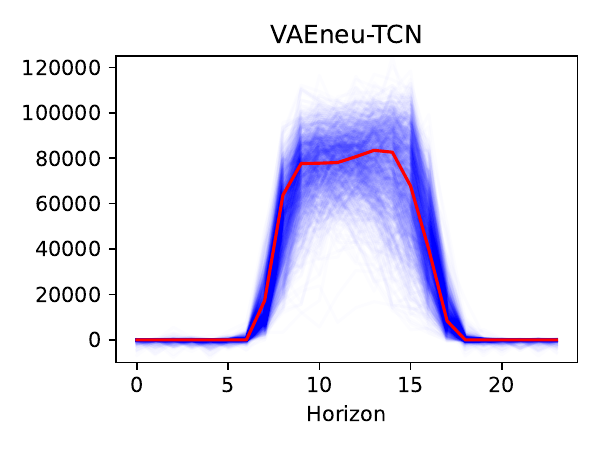}
         \caption{~}
     \end{subfigure}
     \hfill
     \begin{subfigure}{0.32\textwidth}
         \centering
         \includegraphics[width=\textwidth]{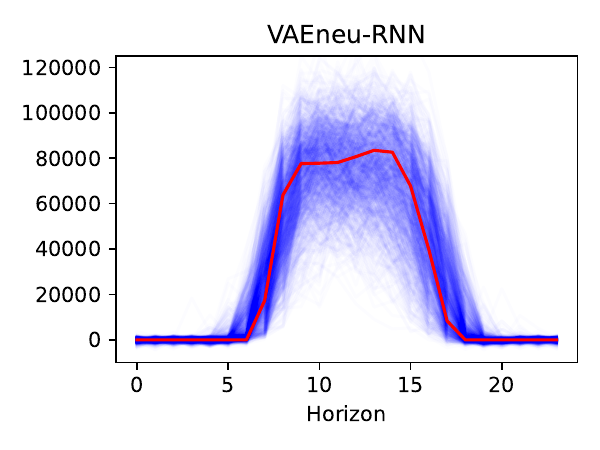}
         \caption{Solar-4-Seconds Dataset}
     \end{subfigure}
     \hfill
    \begin{subfigure}{0.32\textwidth}
         \centering
         
         \includegraphics[width=\textwidth]{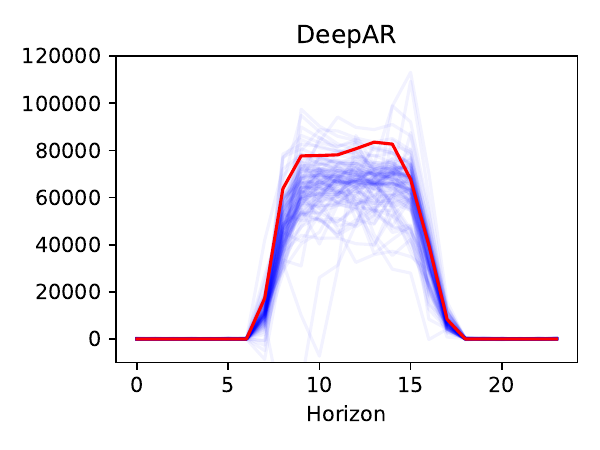}
         \caption{~}
     \end{subfigure}
    \vfill
    \begin{subfigure}{0.32\textwidth}
         \centering
         \includegraphics[width=\textwidth]{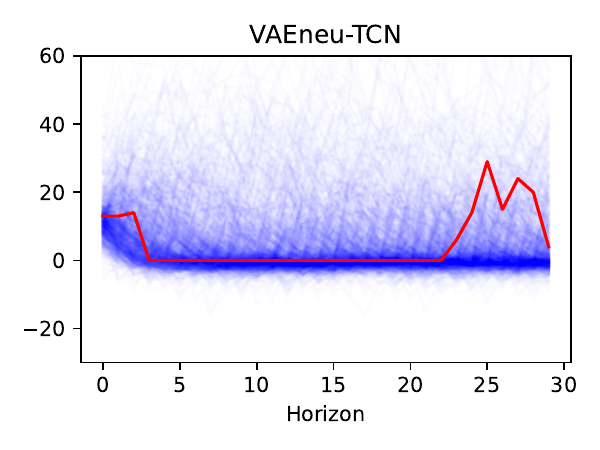}
         \caption{~}
     \end{subfigure}
     \hfill
     \begin{subfigure}{0.32\textwidth}
         \centering
         \includegraphics[width=\textwidth]{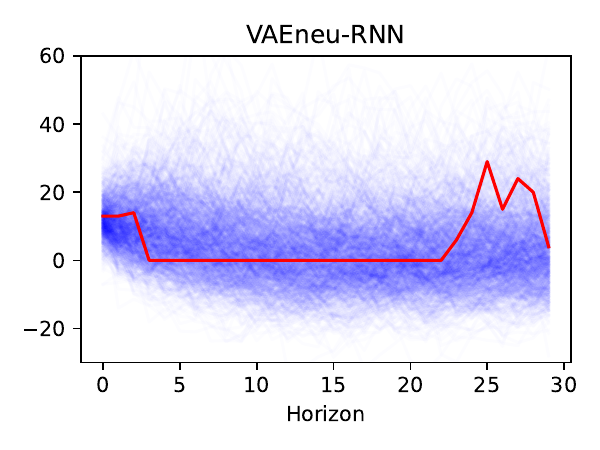}
         \caption{Sunspot Dataset}
     \end{subfigure}
     \hfill
    \begin{subfigure}{0.32\textwidth}
         \centering
         
         \includegraphics[width=\textwidth]{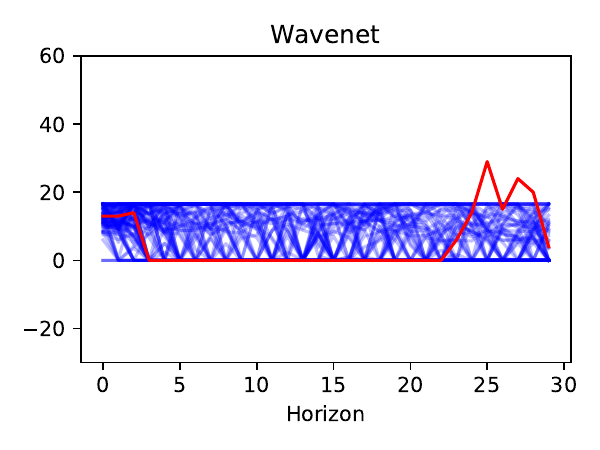}
         \caption{~}
     \end{subfigure}
     
     \vfill
    
    \begin{subfigure}{0.32\textwidth}
         \centering
         \includegraphics[width=\textwidth]{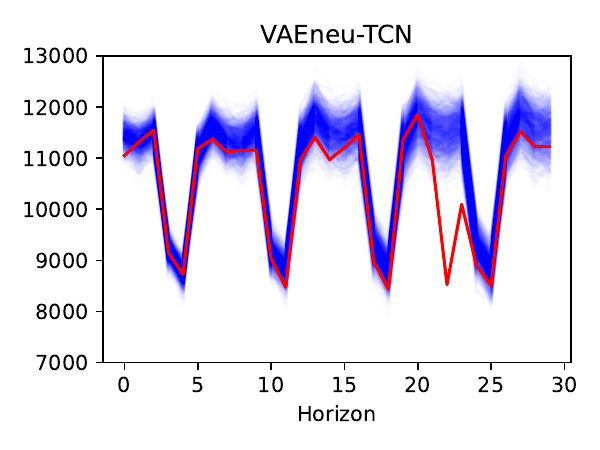}
         \caption{~}
     \end{subfigure}
     \hfill
     \begin{subfigure}{0.32\textwidth}
         \centering
         \includegraphics[width=\textwidth]{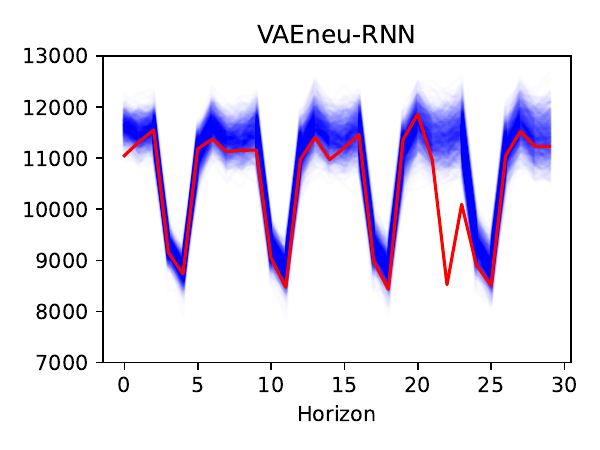}
         \caption{US Birth Dataset}
     \end{subfigure}
     \hfill
    \begin{subfigure}{0.32\textwidth}
         \centering
         
         \includegraphics[width=\textwidth]{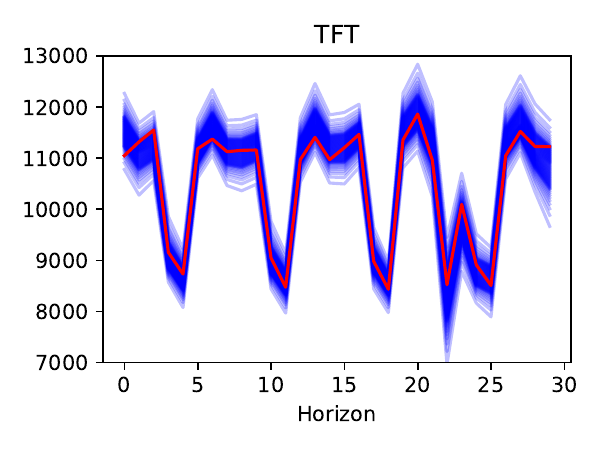}
         \caption{~}
     \end{subfigure}

     \vfill
    
    \begin{subfigure}{0.32\textwidth}
         \centering
         \includegraphics[width=\textwidth]{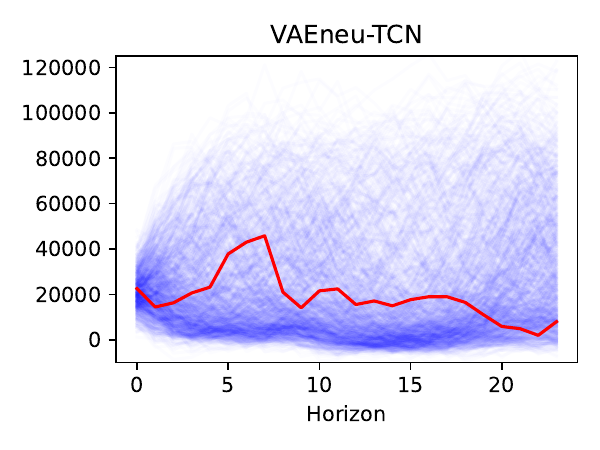}
         \caption{~}
     \end{subfigure}
     \hfill
     \begin{subfigure}{0.32\textwidth}
         \centering
         \includegraphics[width=\textwidth]{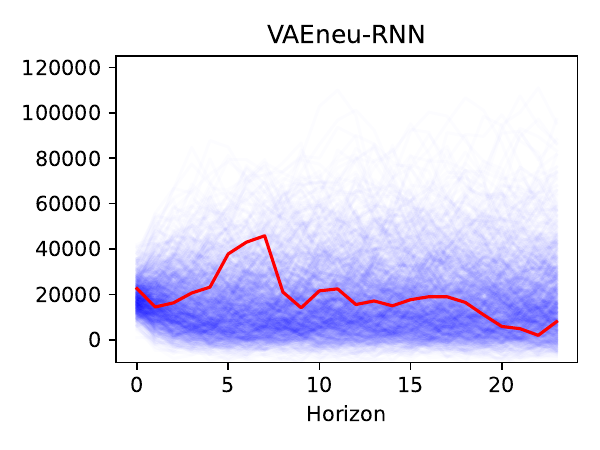}
         \caption{Wind-4-Seconds Dataset}
     \end{subfigure}
     \hfill
    \begin{subfigure}{0.32\textwidth}
         \centering
         
         \includegraphics[width=\textwidth]{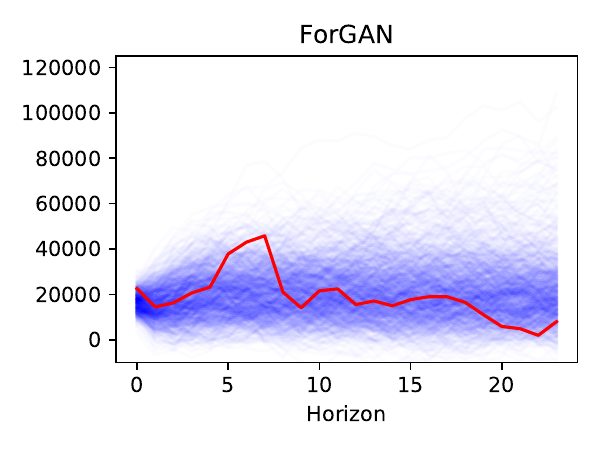}
         \caption{~}
     \end{subfigure}

        \caption{Sample of the forecast for VAEneu models alongside the best baseline model.}
        \label{fig:sample_appendix3}
\end{figure*}

Figures~\ref{fig:sample_appendix1},~\ref{fig:sample_appendix2}, and \ref{fig:sample_appendix3} present a sample forecast for all datasets from VAEneu models and the best baselines for each dataset.

\end{appendices}
\end{document}